%% file: main.tex
\documentclass{article}

\usepackage{microtype}
\usepackage{graphicx}
\usepackage{subfigure}
\usepackage{booktabs} 
\usepackage{wrapfig} 
\usepackage{hyperref}

\usepackage[accepted]{icml2024}

\usepackage{amsmath}
\usepackage{amssymb}
\usepackage{mathtools}
\usepackage{amsthm}
\usepackage{graphicx}
\usepackage{amsmath}
\usepackage{arydshln}
\usepackage{amssymb}
\usepackage{booktabs}
\usepackage{bbm}
\usepackage[utf8]{inputenc} 
\usepackage{bbm, dsfont}
\usepackage{mathtools}
\usepackage{amsthm}
\usepackage[normalem]{ulem}
\usepackage{subcaption}
\usepackage{xcolor,colortbl}
\usepackage{multirow}
\usepackage{pifont}
\usepackage{cutwin}
\usepackage{caption}
\usepackage{enumitem}
\usepackage{wrapfig}
\usepackage{microtype}
\usepackage{bm}
\usepackage{eqparbox}
\usepackage{makecell}
\usepackage{threeparttable}
\usepackage{adjustbox}

\definecolor{Red}{rgb}{0.6,0,0}
\definecolor{Blue}{rgb}{0,0,0.8}
\definecolor{Green}{rgb}{0,0.4,0.7}
\definecolor{airforceblue}{rgb}{0.36, 0.54, 0.66}
\definecolor{ao(english)}{rgb}{0.0, 0.5, 0.0}
\definecolor{azure(colorwheel)}{rgb}{0.0, 0.5, 1.0}
\definecolor{crimson}{rgb}{0.86, 0.08, 0.24}
\definecolor{darkcerulean}{rgb}{0.03, 0.27, 0.49}
\definecolor{cobalt}{rgb}{0.0, 0.28, 0.67}
\definecolor{rosegold}{rgb}{0.72, 0.43, 0.47}
\definecolor{orange-red}{rgb}{1.0, 0.27, 0.0}
\definecolor{mountainmeadow}{rgb}{0.19, 0.73, 0.56}
\definecolor{malachite}{rgb}{0.04, 0.85, 0.32}
\definecolor{darkblue}{rgb}{0.0, 0.0, 0.55}
\definecolor{customblue}{rgb}{0.2, 0.35, 0.8}
\definecolor{beaublue}{rgb}{0.74, 0.83, 0.9}
\definecolor{ggr}{gray}{0.92}
\newcolumntype{a}{>{\columncolor{ggr}}c}
\definecolor{g}{HTML}{EEFFDD}
\definecolor{gg}{HTML}{DDEECC}
\definecolor{ggg}{HTML}{CCDDBB}

\hypersetup{colorlinks=true}
\hypersetup{linktoc=all}
\hypersetup{citecolor=mountainmeadow}
\hypersetup{linkcolor=crimson}
\hypersetup{urlcolor=darkblue}
\usepackage[all]{hypcap}

\usepackage[nameinlink]{cleveref}
\Crefname{section}{Sec.}{Secs.}
\Crefname{algorithm}{Algo.}{Algos.}
\Crefname{table}{Tab.}{Tabs.}
\Crefname{figure}{Fig.}{Figs.}
\Crefname{appendix}{Sec.}{Secs.}

\theoremstyle{plain}

\theoremstyle{definition}

\theoremstyle{remark}

\usepackage[textsize=tiny]{todonotes}

\icmltitlerunning{BECoTTA: Input-dependent Online Blending of Experts for Continual Test-time Adaptation}

\begin{document}

\twocolumn[
\icmltitle{BECoTTA: Input-dependent Online Blending of Experts \\ for Continual Test-time Adaptation}



\icmlsetsymbol{equal}{*}

\begin{icmlauthorlist}
\icmlauthor{Daeun Lee}{equal,yyy}
\icmlauthor{Jaehong Yoon}{equal,xxx}
\icmlauthor{Sung Ju Hwang}{comp,sch}

\end{icmlauthorlist}

\icmlcorrespondingauthor{Daeun Lee}{goodgpt@korea.ac.kr}
\icmlcorrespondingauthor{Jaehong Yoon}{jhyoon@cs.unc.edu}
\icmlcorrespondingauthor{Sung Ju Hwang}{sjhwang82@kaist.ac.kr}

\icmlaffiliation{yyy}{Statistics, Korea University}
\icmlaffiliation{xxx}{Computer Science, UNC-Chapel Hill}
\icmlaffiliation{comp}{Korea Advanced Institute of Science and Technology}
\icmlaffiliation{sch}{DeepAuto}


\icmlkeywords{Machine Learning, ICML}

\vskip 0.3in 
]


\printAffiliationsAndNotice{\icmlEqualContribution} 

\input{tex/01_abstract}

\input{tex/02_intro}

\input{tex/03_related}
\input{tex/05_methods}
\input{tex/06_experiments}
\input{tex/07_conclusion}

\bibliography{main}
\bibliographystyle{icml2024}

\newpage

\appendix
\onecolumn
\input{tex/Appendix}


\end{document}

%% file: tex/01_abstract.tex
\begin{abstract}
Continual Test Time Adaptation (CTTA) is required to adapt efficiently to continuous unseen domains while retaining previously learned knowledge. 
However, despite the progress of CTTA, it is still challenging to deploy the model with improved forgetting-adaptation trade-offs and efficiency.  
In addition, current CTTA scenarios assume only the disjoint situation, even though real-world domains are seamlessly changed. 
To address these challenges, this paper proposes \textit{BECoTTA}, an input-dependent and efficient modular framework for CTTA. 
We propose Mixture-of-Domain Low-rank Experts (MoDE) that contains two core components: \textit{(i) Domain-Adaptive Routing}, which helps to selectively capture the domain-adaptive knowledge with multiple domain routers, and \textit{(ii) Domain-Expert Synergy Loss} to maximize the dependency between each domain and expert. 
We validate that our method outperforms multiple CTTA scenarios, including disjoint and gradual domain shits, while only requiring $\sim$98\% fewer trainable parameters. 
We also provide analyses of our method, including the construction of experts, the effect of domain-adaptive experts, and visualizations. Project Page: \href{https://becotta-ctta.github.io/}{\textcolor{magenta}{\textit{https://becotta-ctta.github.io/}}}.
\end{abstract}

%% file: tex/02_intro.tex
\section{Introduction}

\begin{figure}[!]
    \centering
    \includegraphics[width=1.0\linewidth]{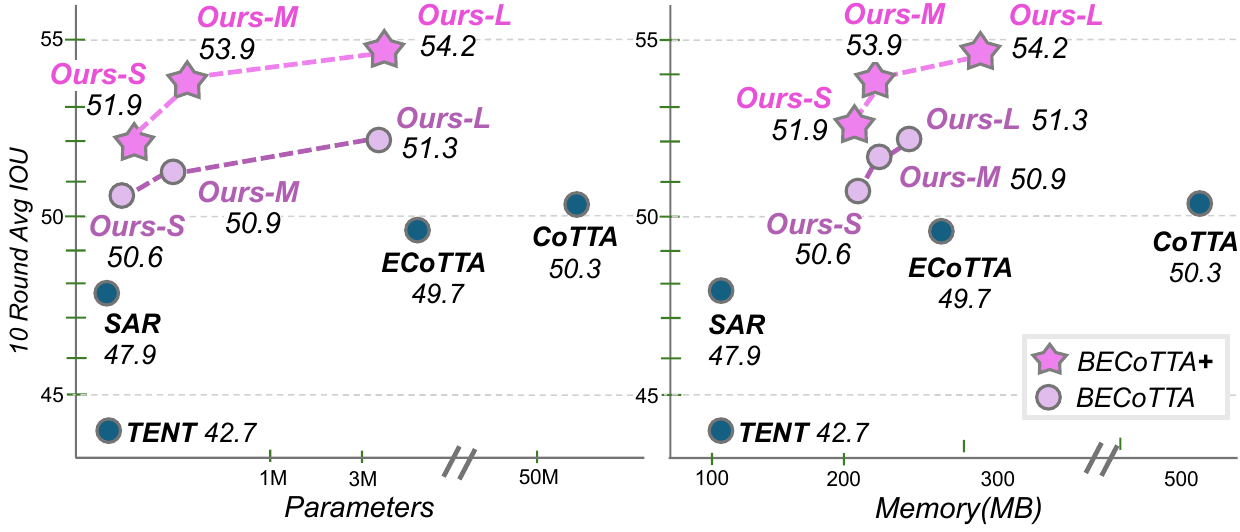}
    \vspace{-.5cm}
    \caption{
    BECoTTA and BECoTTA+ achieve superior 10-round average IoU and parameter/memory efficiency against strong CTTA baselines on the CDS-\textit{hard} scenario.} 
    \label{fig:main_performance}
    \vspace{-.2cm}
\end{figure}

Test-Time Adaptation (TTA)~\cite{tent,sar,cotta,lim2023ttn, wisdom} is a challenging task that aims to adapt the pre-trained model to new, unseen data at the time of inference, where the data distribution significantly differs from that of the source dataset. TTA approaches have become popular since they address the critical challenge of model robustness and flexibility in the face of new data.

\begin{figure*}[h]
    \centering
    {\includegraphics[width=0.88\textwidth]{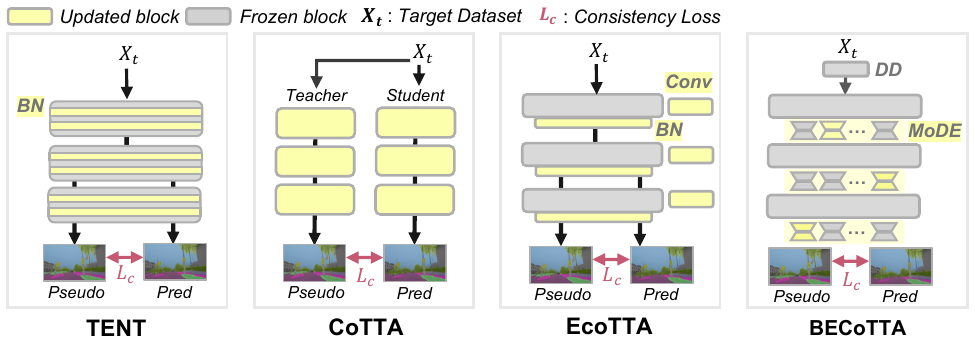}}
    \caption{\textbf{Comparison of TTA process with other SoTA models.} We compare the existing models~\cite{tent, cotta, ecotta} and denote activated modules as yellow during CTTA process. In particular, CoTTA adopts the mean-teacher architecture and updates the entire model. 
    TENT~\cite{tent} and EcoTTA~\cite{ecotta} update only a few parameter-efficient modules in the model. However, they achieve suboptimal performance with forgetting. 
    Meanwhile, our BECoTTA updates only MoDE layers for efficient and rapid adaptation while preserving previous knowledge.}
    \label{fig:comparison}
    \vspace{-0.15in}
\end{figure*}

Beyond the isolated transferability of traditional TTA approaches on the stationary target domain, Continual Test-Time Adaptation (CTTA)~\cite{sar,ecotta,wisdom} has been increasingly investigated in recent years, whose goal is to \emph{continuously} adapt to multiple unseen domains arriving in sequence.
Solving the problem of CTTA is crucial because it is closely related to real-world scenarios. 
For example, let us assume that a vision model in an autonomous vehicle is designed to understand road conditions and objects, including pedestrians, vehicles, traffic signs, etc. 
The agent will encounter different environments over time, depending on changes in weather, time of day, and location. 
Then, the model should rapidly and continuously adapt to these unseen environments while retaining the domain knowledge learned during adaptation as it may re-encounter prior domains in the future.   
 
Therefore, continual TTA approaches need to address the following key challenges: (i) \textit{forgetting-adaptation trade-off}: retaining previous domain knowledge while learning new domains often limits the model's plasticity, hindering its ability to learn and adapt to new data, and (ii) \textit{computational efficiency}: since CTTA models are often assumed to be embedded in edge devices, efficient adaptation is significant. However, existing CTTA methods overlook computational efficiency by updating heavy teacher and student models~\cite{cotta} or achieved suboptimal convergence due to updating only a few parts of modules~\cite{tent, sar, vdp, dept}.

To tackle these critical issues, this paper proposes \textit{Input-dependent Online \textbf{B}lending of \textbf{E}xperts for \textbf{Co}ntinual \textbf{T}est-\textbf{T}ime \textbf{A}daptation (\textbf{BECoTTA})} by introducing a surprisingly efficient yet effective module, named \textit{Mixture-of-Domain Low-rank Experts (MoDE)}, atop each backbone block. 
Our BECoTTA method consists of two key components: \textit{(i) Domain Adaptive Routing} and \textit{(ii) Domain-Expert Synergy Loss}. We first propose Domain Adaptive Routing that aims to cluster lightweight low-rank experts (i.e., MoDE modules) with relevant domain knowledge. Next, based on the assignment of domain adaptive routers, we propose Domain-Expert Synergy Loss to maximize mutual information between each domain and its corresponding expert. In the end, we facilitate cooperation and specialization among domain experts by ensuring strong dependencies.
Our modular design allows for selective updates of multiple domain experts, ensuring the transfer of knowledge for each specific domain while preserving previously acquired knowledge.
This approach also significantly improves memory and parameter efficiency through sparse updates.

Furthermore, existing CTTA scenarios assume a \textit{disjoint} change of test domains, where the model encounters a static domain per time step, but do not consider a \emph{gradual shift of domains}, which is more common in the real world (e.g., seamless weather change like cloudy $\rightarrow$ rainy or afternoon $\rightarrow$ night). 
To further consider this realistic scenario, we additionally propose 
\textit{Continual Gradual Shifts (CGS)} benchmark for CTTA, where the domain gradually shifts over time based on the domain-dependent sampling distribution, as illustrated in \Cref{fig:overall} top left. 
This scenario is more advanced than an existing CTTA problem as it demands the model to appropriately adapt each of the input instances, without relying on any implicit guidance from the dominant domain over a given time interval.

We compare our proposed method with strong baselines, including SAR~\cite{sar}, DePT~\cite{dept}, VDP~\cite{vdp}, and EcoTTA~\cite{ecotta}, on multiple CTTA scenarios and our suggested CGS benchmark.
Our BECoTTA achieves \textbf{+2.1\%\textit{p} and +1.7\%\textit{p} IoU enhancement} respectively on \textit{CDS-Hard} and \textit{CDS-Easy} scenarios, by \textbf{utilizing \textbf{95\%} and \textbf{98\%} fewer parameters} used by CoTTA~\cite{cotta}. 
Furthermore, we propose BECoTTA+, which is initialized by the source augmentation dataset. BECoTTA+ shows increased performance by \textbf{+16.8\%\textit{p} compared to EcoTTA (initialized fairly)}, utilizing a similar number of parameters on the \textit{CDS-Hard} scenario. 

We summarize our contributions as threefold: 
\begin{itemize}
\item We propose an efficient yet powerful CTTA method, named \textit{BECoTTA}, which adapts to new domains effectively with minimal forgetting of the past domain knowledge, by transferring only beneficial representations from relevant experts.

\item We introduce a new realistic CTTA benchmark, \textit{Continual Gradual Shifts (CGS)} where the domain gradually shifts over time based on domain-dependent continuous sampling probabilities.

\item We validate our BECoTTA on various driving scenarios, including three CTTA and one domain generalization, demonstrating the efficacy and efficiency against strong baselines, including TENT, EcoTTA, and SAR.

\end{itemize}

%% file: tex/03_related.tex
\section{Related Works}

\paragraph{Continual Test-Time Adaptation.}
Continual Test-Time Adaptation (CTTA)~\cite{cotta,vdp,sar,ecotta} assumes that target domains are not fixed but change continuously in an online manner. 
TENT~\cite{tent} is one of the pioneering works, which activates only BatchNorm layers to update trainable affine transform parameters. 
CoTTA~\cite{cotta} introduces a teacher-student framework, generating pseudo-labels from the teacher model, and updating it using consistency loss. 
EcoTTA~\cite{ecotta} utilizes meta-networks and self-distilled regularization while considering memory efficiency. 
DePT~\cite{dept} integrates visual prompts to efficiently adapt target domains and bootstraps the source representation.   
However, existing methods often suffer from subordinate convergence, as they rely on a shared architecture to adapt the test data without considering the correlation between different domains. 
On the other hand, our BECoTTA introduces a modularized MoE-based architecture where each expert captures domain-adaptive knowledge, and the model transfers only a few related experts for the adaptation of a new domain.

Moreover, recent works~\cite{ecotta, eata, lim2023ttn, choi2022improving, liu2021ttt++, adachi2022covariance, jung2023cafa, lee2023tta} allow for a slight warm-up using the source dataset before deploying the model to the CTTA scenario. 
In particular, TTA-COPE~\cite{lee2023tta} performs pretraining with labeled source datasets in a supervised manner. EcoTTA~\cite{ecotta} also allows warm-up to initialize their meta-network. 
Note that these methods still assume \textit{source-free} training during test-time adaptation, which means that this setup adheres to the assumptions of CTTA.

\paragraph{Mixture-of-Experts.}
Mixture-of-Experts (MoE)~\cite{moe1, moe2, thor, adamix} introduces $N$ parallel experts consisting of the feedforward network with router modules and sparsely activates a few experts based on their sampling policies. Adamix~\cite{adamix} introduces efficient fine-tuning with Mixture-of-Adapters to learn multiple views from different experts. THOR~\cite{thor} proposes a new stochastic routing function to prevent inefficiency with routers. Similarly, Meta DMoE~\cite{zhong2022meta} adopts the MoE architecture as a teacher model and distills their knowledge to unlabeled target domains, but does not consider continuous adaptation. In short, to the best of our knowledge, the feasibility of MoE structures is underestimated in the field of CTTA.


\paragraph{Blurry Scenario in Continual Learning.}
Recently, a few continual learning appraoches~\cite{blurry1, bang2021rainbow, blurry2, blurry3} have discussed \textit{Blurry Continual Learning (Blurry-CL)} to better reflect real-world scenarios, beyond the standard CL setting.
Blurry-CL assumes that for each sequential task, a majority class exists, and other classes outside the majority class may also overlap and appear. 
The most renowned scenario setup is a \textit{Blurry-M}~\cite{blurry3}, where the majority class occupies 100-M\%, and the remaining classes are randomly composed of M\%. 
Although this benchmark handles an overlapping situation, it may not cover practical situations where the domain evolves gradually in CTTA. 
Therefore, we propose a new benchmark that simulates real-world continual TTA scenarios with a gradual change of domains over time.

%% file: tex/05_methods.tex
\section{Input-dependent Online Blending of Experts for Continual Test-time Adaptation} 

We first define the problem statement for Continual Test-Time Adaptation (CTTA) in~\Cref{sec:subsec:problem}. Next, we introduce our proposed CTTA method, BeCoTTA, containing Mixture of Domain low-rank Experts (MoDE) and domain-expert synergy loss in~\Cref{sec:subsec:dai,sec:subsec:mode}. Then, we describe the overall optimization process during CTTA in~\Cref{sec:subsec:adaptation}.

\subsection{Problem Statement}\label{sec:subsec:problem}
Continual Test-time Adaptation (CTTA) aims to adapt the pre-trained source model $\bm{f_s}$ to continuously changing target domains, formulated as a task sequence $\bm{X_t}=\{X_t^1, X_t^2, .. X_t^c, \cdots\}$. 
The main assumption of CTTA includes that (i) we should not access the source dataset \textit{after} deploying the model to the test time, and (ii) adaptation needs to be done online and in an unsupervised manner. 
For semantic segmentation tasks, CTTA aims to predict the softmax output $\hat{y_c} = f_c(\bm{x}_t^c)$ in the target domain $c$. 
$\bm{x}_t^c$ is sampled from $X_t^c$, which will be represented by $\bm{x}$ in the following sections for brevity.

\begin{figure*}[h]
    \centering
    {\includegraphics[width=1.0\textwidth]{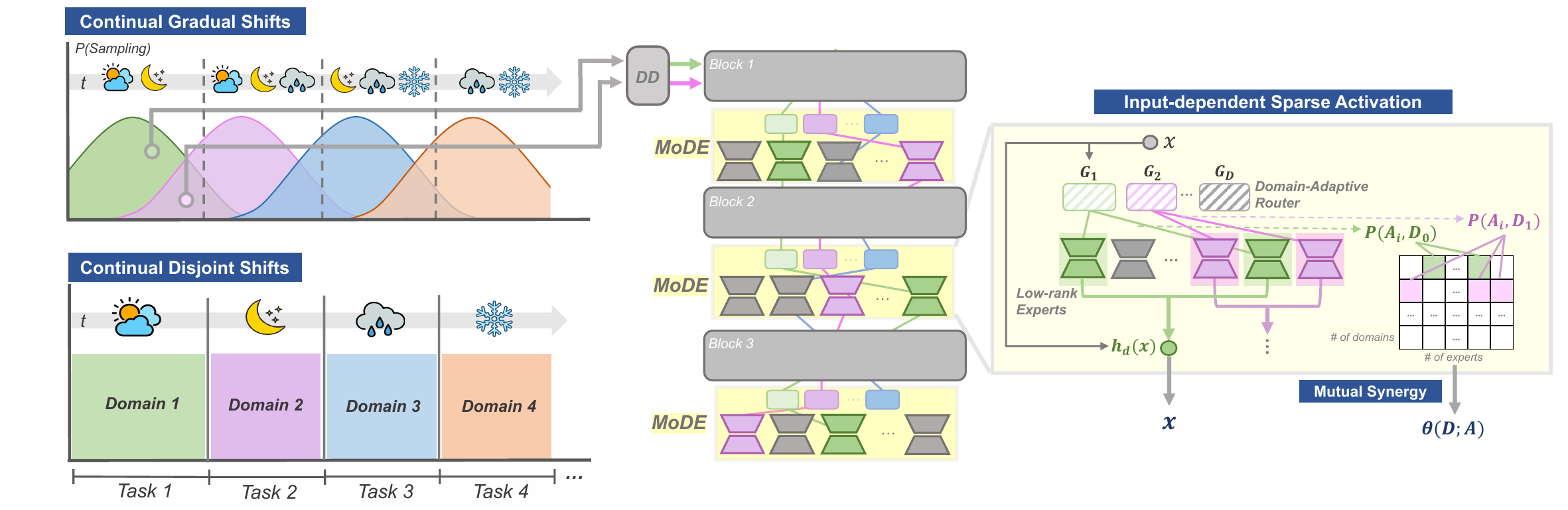}}
    \caption{\textbf{The overview of BECoTTA.} We propose a novel CTTA framework for dynamic real-world scenarios, including disjoint and gradual shifts of domains. When the model receives a target domain input $\bm{x_t}$ at timestep $t$, the Domain Discriminator (DD) first estimates a pseudo-domain label $d$. Based on estimated pseudo-labels, the domain router $G_d$ processes the input to specific experts containing domain-specific information by minimizing \textit{Domain-Expert Synergy Loss} $\Theta(D;A)$. Finally, we obtain a domain-adaptive representation $h_d(\bm{x})$, addressing downstream tasks in test-time.}
    \label{fig:overall} 
\end{figure*}

\subsection{Domain-Augmented Initialization.}\label{sec:subsec:dai}
\textbf{Source Domain Augmentation (SDA).} Most CTTA methods use a pre-trained frozen backbone, which contains domain bias from the source domain. Due to the predominance of the source domain, this bias impedes the effective transfer of domain-adaptive knowledge in continuous scenarios.
To mitigate this limitation, we define $D$ proxy domains (e.g., brightness, darkness, blur, etc.) and augment the source dataset to proxy domains, similar to EcoTTA~\cite{ecotta}. 
For this Source Domain Augmentation (SDA), we utilize pre-trained style-transfer~\cite{jiang2020tsit} or simple transformations~\cite{Albumentations}. 
Through SDA, we acquire domain-specific knowledge before deploying TTA. This process is done only \textit{once} when constructing the source dataset.

\textbf{Robustness to SDA.} We emphasize that these pre-defined domains \textbf{do not need to match} CTTA target domains. 
The primary role of SDA is to differentiate routers so that the model can aggregate different visual features during the continual TTA phases. 
Our BECoTTA+ is able to update relevant MoDE modules with respect to the inputs and consistently achieves competitive performance even when the SDA and target domains are disjoint. (Please refer to \Cref{table:wad_ablation} and Appendix (\Cref{append:data}) for more details.) For a fair comparison with other CTTA methods, we also verify that \textit{random} and \textit{source-domain-only} initializations work well with BECoTTA. More details are provided in Appendix (\Cref{tab:init}).

\subsection{Mixture-of-Domain Low-rank Experts (MoDE)}\label{sec:subsec:mode}

We now introduce our new CTTA approach to efficiently capture the domain-adaptive representation via cooperation and specialization of multiple experts, dubbed \textit{Input-dependent Online \textbf{B}lending of \textbf{E}xperts for \textbf{Co}ntinual \textbf{T}est-\textbf{T}ime \textbf{A}daptation (\textbf{BECoTTA})}. 
Our proposed BECoTTA employs \textit{Mixture of Domain low-rank Experts (MoDE)} layers at the top of each block in the pre-trained ViT backbone.

\textbf{The design of Low-rank Experts.}
For the efficient process during CTTA, we adopt the Sparse Mixture-of-Experts (SMoE) module with a top-k routing policy~\cite{moe1}. 
Each MoE layer consists of the router $G$ and a set of $N$ lightweight experts, $A_1$, $A_2$, ..., $A_N$, where each $A_i$ is parameterized by $\bm{W}_i^{down} \in\mathbb{R}^{dim\times r}$ and $\bm{W}_i^{up}\in\mathbb{R}^{r\times dim}$. Here, $r$ denotes the rank, and $dim$ denotes the embedding dimension of each ViT block.
If $A_i$ is activated, it maps the input $\bm{x}$ into the low-dimensional space through the projection operation with $\bm{W}_i^{down}$.
Next, after regularizing the features with a non-linear activation function $\sigma(\cdot)$, it recovers the features to the original dimension using $\bm{W}_i^{up}$: 
\begin{equation}
A_i = \sigma(\bm{x}\bm{W}_i^{down})\bm{W}_i^{up}.
\end{equation}

\textbf{Domain-Adaptive Routing.} 
Since each domain contains different key features, transferring them to other domains is not always advantageous. 
However, recent CTTA approaches, such as TENT~\cite{tent}, SAR~\cite{sar}, and EcoTTA~\cite{ecotta}, continuously adapt to new domains by updating trainable parameters in a domain-agnostic manner. This means that they update the equivalent set of parameters for adapting a variety of different domains over time, which restricts the ability to learn fine-grained features for each domain due to the negative interference from irrelevant domain knowledge.
In addition, adjusting all parameters for new domains causes the model to forget the past domain information, struggling to retain domain representations learned before when encountering the same or similar domains again.

Therefore, as shown in~\Cref{fig:overall}, we introduce $D$ independent domain-wise routers $G_1$, $G_2$..$G_D$ to loosely cluster experts of the model with similar domain knowledge by selecting $K$ experts per layer. 
We note that our modular architecture containing multiple parameter-efficient experts allows the model to efficiently yet effectively capture domain-adaptive representations while avoiding negative interference from less relevant features and preventing unintentional shifts of previously learned domains. 
Each router $G_d$ for domain $d$ is parameterized by $\bm{W}^g_d\in\mathbb{R}^{dim\times N}$ and $\bm{W}^{noise}_d\in\mathbb{R}^{dim\times N}$, and operates as follows: 
\begin{equation}
P_d(\bm{x})=\bm{x}\bm{W}^g_d+N\left(0,1\right)\cdot\text{\text{Softplus}}\left(\bm{x}\bm{W}^{noise}_d\right),
\end{equation}
\begin{equation}
G_d(\bm{x})=\text{Softmax}\left(TopK\left( P_d(\bm{x}) \right) \right).
\end{equation}
Based on the $G_d(\bm{x})$, we \textit{selectively} update the activated experts associated with the specific domain, inherently isolating them from irrelevant domain knowledge while adapting new ones. 
In the end, the output of the MoDE layer $h_d(x)$ aggregates the domain-adaptive features as follows:
\begin{equation}
h_d(\bm{x})=\sum_{i=1}^NG_d^i(\bm{x})\cdot A_i.
\end{equation}
This trainable clustering approach allows the model to activate its own set of experts, who are specialized in specific domain knowledge. Moreover, our multi-router-based design accelerates adaptation to the current domain, avoiding interference from knowledge transfer of unrelated domain features. 
Finally, we perform the skip connection operation with the original input $\bm{x}$: $\bm{x} \leftarrow \bm{x} + h_d(\bm{x})$. 

\textbf{Maximizing Domain-Expert Synergy.}  
In cases where some domains share similar visual contexts (e.g., \textit{snow} and \textit{fog}), collaboration between domain experts can be beneficial. On the other hand, for unique scenes like \textit{night}, it is advantageous to isolate domain features from others. That is, ensuring strong interdependence among various domains and experts is essential. To this end, we propose Domain-Expert Synergy loss based on the output from domain-adaptive routers. 
Let us consider $G_d^i(\bm{x})$ as the \textit{assignment weight} with specific domain $d$ of the $i$-th expert $A_i$, then $P(A_i|d)$ is obtained from all the experts and domains in each MoDE layer. Then, we calculate $P(A_i, d)$ using Bayes' theorem: 
\begin{equation}
P(A_i, d) = P(A_i|d) \cdot P(d),
\end{equation}
where $P(d)$ represents the frequency of occurrence in domain $d$. Since it is infeasible to define $P(d)$ in most real-world scenarios, we assume the uniform distribution over $P(d)$. 
Next, to measure and maximize the mutual dependency among domains and experts, we adopt the probability modeling as a double sum:
\begin{equation}
\Theta(D;A)=\sum_d^D\sum_i^NP(A_i, d)\cdot \log \frac{P(A_i, d)}{P(A_i)P(d)}.
\end{equation}
Maximizing $P(A_i, d) \log P(A_i, d)$ leads the model to obtain a sharper conditional distribution of $P(A_i|d)$, facilitating the dependency between domains and experts. That is, our domain-adaptive experts can be further specialized in their respective domains and collaborate with others who share their domain knowledge. 

\subsection{Continual Test-time Adaptation Process}    \label{sec:subsec:adaptation}       

\textbf{Model Initialization.}
Following recent trends in CTTA~\cite{ecotta, eata, lim2023ttn, choi2022improving, liu2021ttt++, adachi2022covariance, jung2023cafa} and for a fair comparison, we perform a short pre-training for trainable parameters in models before deploying them to the CTTA problems.
We initialize our BECoTTA with \textit{three} different manners: (i) \textit{random},  (ii) \textit{source-domain-only} (w/o SDA, \texttt{BECoTTA}), and (iii) \textit{domain-augmented} (w/ SDA, \texttt{BECoTTA+}) initialization.
For \textit{(i) random} initialization, we randomly initialize MoDE layer weights. 
For \textit{(ii) source-domain-only} initialization, following EcoTTA~\cite{ecotta}, we initialize MoDE layer weights using the source domain. 
Note that this initialization strategy regards the fair comparison in the CTTA literature.
For \textit{(iii) domain-augmented initialization}, we first build $D$ domains using SDA. Next, we introduce a Domain Discriminator (DD) as the auxiliary head. It consists of lightweight CNN layers and is trained to classify the pre-defined $D$ domains. This helps the model distinguish between different domains and classify each test-time image input accordingly during test-time adaptation on a sequence of unseen domains. Then, we update MoDE layers and DD for only a small number of epochs.    
The total initialization loss $L_{init}$ is formulated below. Except for the original cross-entropy loss $L_{seg}$ for semantic segmentation, we include the cross-entropy loss $L_{disc}$ for DD, and the domain-expert synergy loss $\Theta(D;A)$: 
\begin{equation}
L_{init} = L_{seg} + \lambda_{disc}L_{disc} - \lambda_s\Theta(D;A)
\end{equation}
where $\lambda_{disc}$, $\lambda_s$ denotes the balance term for each loss.

\textbf{Source-free CTTA with MoDE.} 
Building upon the initialized MoDE, we deploy our BECoTTA to the continual target domains $\bm{X_t}$.
Note that \textbf{we do not access any source dataset after deployment}, maintaining the \textit{source-free} manner in the test time as other prior works.  
In the source-free TTA, we only activated MoDE layers to transfer the target domain knowledge efficiently.  
Utilizing the frozen DD trained at initialization, we obtain the pseudo-domain label $d$ for each domain-agnostic target image $x_t^c$. 
Afterward, according to $d$, we initially assign the domain-wise router and proceed with the aggregation of domain-adaptive experts. 
This approach ensures that our BECoTTA+ maintains input dependency, even within unseen test-time domains.

\input{table/wo_ours}

Following preliminary works~\cite{tent, ecotta, eata}, we adopt the entropy minimization using $H(\hat{y_c}) = -\sum_{}^{}p(\hat{y_c}) \cdot \log p(\hat{y_c})$.
To avoid forgetting and error accumulation, we perform entropy filtering based on the confidence of the pseudo-labels.
Therefore, the entropy-based loss $L_{tta}$ is as follows:
\begin{equation}
L_{tta}=\mathds{1}_{\{{H(\hat{y_c})<\kappa}\}} \cdot H(\hat{y_c}),
\end{equation}
where $\hat{y_c}$ is the output prediction in the current target domain stage $c$, $\kappa$ is the pre-defined entropy threshold, and $\mathds{1}\{\cdot\}$ denotes an indicator function.

%% file: table/wo_ours.tex
\begin{table*}[h]
\centering
\renewcommand{\arraystretch}{1.25}
\caption{\textbf{Results on \textit{CDS-Hard (imbalanced weather \& area shifts)}.} We devise a novel scenario encompassing imbalanced weather and area shifts. We present performance results for both \textit{w/o SDA} and \textit{w/ SDA} across the overall baselines. We report $S$, $M$, and $L$ versions for our BECoTTA based on the number of parameters. }
\setlength{\tabcolsep}{.2em}
\resizebox{\linewidth}{!}{%
\begin{tabular}{clc|cccccc|cccccc|c|c}
\toprule

\multicolumn{3}{c|}{Round} & \multicolumn{6}{c|}{\textbf{1}} & \multicolumn{6}{c|}{\textbf{10}} & \multirow{2}{*}{$\Delta$} &  Parameter \\ \cline{1-15}

\multicolumn{1}{c|}{Init} & \multicolumn{1}{c|}{Method}  & Activated & B-Clear & A-Fog & A-Night & A-Snow & \multicolumn{1}{c|}{B-Overcast} & Mean & \multicolumn{1}{l}{B-clear} & \multicolumn{1}{l}{A-Fog} & \multicolumn{1}{l}{A-Night} & \multicolumn{1}{l}{A-Snow} & \multicolumn{1}{l|}{B-Overcast} & \multicolumn{1}{c|}{Mean} & \\ \hline \hline

\multicolumn{1}{l|}{} & \multicolumn{1}{l|}{Source only}  & - &41.0 & 64.4 & 33.4 & 54.3 & \multicolumn{1}{c|}{46.3} & 47.9 & 41.0 & 64.4 & 33.4 & 54.3 & \multicolumn{1}{c|}{46.3} & 47.9 & +0.0 & - \\ 

\multicolumn{1}{l|}{} & \multicolumn{1}{l|}{CoTTA~\cite{cotta}}  & - & 43.3 & 67.3 & 34.8 & 56.9 & \multicolumn{1}{c|}{48.8} & 50.2 & 43.3 & 67.3 & 34.8 & 56.9 & \multicolumn{1}{c|}{48.8} & 50.2 & +0.0  & \textbf{54.72M}\\

\multicolumn{1}{l|}{} & \multicolumn{1}{l|}{TENT~\cite{tent}} & - & 41.1 & 64.9 & 33.2 & 54.3 & \multicolumn{1}{c|}{46.3} & 47.9 & 30.9 & 51.5 & 20.4 & 37.0 & \multicolumn{1}{c|}{33.0} & 34.6 & \textcolor{red}{-13.3} & 0.02M \\
    
\multicolumn{1}{l|}{\textit{w/o SDA}} & \multicolumn{1}{l|}{SAR~\cite{sar}} &  - & 41.0 & 64.5 & 33.4 & 54.5 & \multicolumn{1}{c|}{46.6} & 48.0 & 41.3 & 64.3 & 31.6 & 54.2 & \multicolumn{1}{c|}{46.6} & 47.6 & \textcolor{red}{-0.4}  & 0.02M \\
    
\multicolumn{1}{l|}{} & \multicolumn{1}{l|}{EcoTTA~\cite{ecotta}} & \textit{\textcolor{darkcerulean}{MetaNet}} & 44.1 & 69.6 & 35.3 & 58.2 & \multicolumn{1}{c|}{49.6} & 51.3 & 41.9 & 66.1 & 31.5 & 55.3 & \multicolumn{1}{c|}{46.2} & 48.2 & \textcolor{red}{-3.1} & 3.46M\\ 

\multicolumn{1}{l|}{}& \multicolumn{1}{l|}{\cellcolor{g}BECoTTA (S)} & 
\cellcolor{g}\textit{\textcolor{darkcerulean}{MoDE}} & 
\cellcolor{g}42.9 &
\cellcolor{g}68.5 &
\cellcolor{g}35.0 &
\cellcolor{g}57.2 &
\multicolumn{1}{c|}{\cellcolor{g}47.8} &
\cellcolor{g}50.5 &
\cellcolor{g}43.0 &
\cellcolor{g}68.5 &
\cellcolor{g}35.1 &
\cellcolor{g}57.3 &
\multicolumn{1}{c|}{\cellcolor{g}48.8} &
\cellcolor{g}50.7 &
\cellcolor{g}+0.1 &
\cellcolor{g}0.09M \\

\multicolumn{1}{l|}{}& \multicolumn{1}{l|}{\cellcolor{gg}BECoTTA (M)} & 
\cellcolor{gg}\textit{\textcolor{darkcerulean}{MoDE}} & 
\cellcolor{gg}43.8 &
\cellcolor{gg}68.8 &
\cellcolor{gg}34.9 &
\cellcolor{gg}57.9 &
\multicolumn{1}{c|}{\cellcolor{gg}49.2} & 
\cellcolor{gg}50.9 & 
\cellcolor{gg}43.7 & 
\cellcolor{gg}68.8 & 
\cellcolor{gg}34.5 & 
\cellcolor{gg}57.9 & 
\multicolumn{1}{c|}{\cellcolor{gg}49.2} &
\cellcolor{gg}50.9 &
\cellcolor{gg}+0.0 & 
\cellcolor{gg}0.63M \\

\multicolumn{1}{l|}{}& \multicolumn{1}{l|}{\cellcolor{ggg}BECoTTA (L)} &
\cellcolor{ggg}\textit{\textcolor{darkcerulean}{MoDE}} & 
\cellcolor{ggg}\textbf{43.9} &
\cellcolor{ggg}\textbf{69.1} &
\cellcolor{ggg}\textbf{35.0} &
\cellcolor{ggg}\textbf{58.3} &
\multicolumn{1}{c|}{\cellcolor{ggg}\textbf{50.2}} &
\cellcolor{ggg}\textbf{51.3} &
\cellcolor{ggg}\textbf{44.0} &
\cellcolor{ggg}\textbf{69.1} &
\cellcolor{ggg}\textbf{35.1} &
\cellcolor{ggg}\textbf{58.3} &
\multicolumn{1}{c|}{\cellcolor{ggg}\textbf{50.2}} &
\cellcolor{ggg}\textbf{51.3} &
\cellcolor{ggg}+0.0 &
\cellcolor{ggg}3.16M \\ \hline

\multicolumn{1}{l|}{}  & \multicolumn{1}{l|}{Source only}  & \textit{\textcolor{crimson}{Full}} & 43.6 & 68.7 & 44.5 & 59.0 & \multicolumn{1}{c|}{48.7} & 52.9 & 43.6 & 68.7 & 44.5 & 59.0 & \multicolumn{1}{c|}{48.7} & 52.9 & +0.0 & -\\  

\multicolumn{1}{l|}{} & \multicolumn{1}{l|}{CoTTA~\cite{cotta}} & \textit{\textcolor{crimson}{Full}} & 46.4 & 70.6 & 45.7 & 61.2 & \multicolumn{1}{c|}{51.3} & 55.0 & 46.1 & 70.5 & 45.6 & 61.1 & \multicolumn{1}{c|}{51.2} & 54.9 & \textcolor{red}{-0.1}  & \textbf{54.72M}\\ 

\multicolumn{1}{l|}{} & \multicolumn{1}{l|}{TENT~\cite{tent}} & \textit{\textcolor{crimson}{Full}} & 43.7 & 68.5 & 44.6 & 59.0 & \multicolumn{1}{c|}{48.3} & 52.8 & 35.8 & 57.6 & 33.6 & 44.3 & \multicolumn{1}{c|}{38.8} & 42.0 & \textcolor{red}{-10.8} & 0.02M \\ 

\multicolumn{1}{l|}{\textit{w/ SDA}} & \multicolumn{1}{l|}{SAR~\cite{sar}}  & \textit{\textcolor{crimson}{Full}} & 43.6 & 68.6 & 44.5 & 59.1 & \multicolumn{1}{c|}{48.7} & 52.9 & 43.4 & 67.4 & 42.2 & 58.1 & \multicolumn{1}{c|}{47.6} & 51.9 & \textcolor{red}{-1.0} & 0.02M \\ 

\multicolumn{1}{l|}{} & \multicolumn{1}{l|}{EcoTTA~\cite{ecotta}} & \textit{\textcolor{darkcerulean}{MetaNet}} & 44.6 & 70.2 & 41.6 & 58.0 & \multicolumn{1}{c|}{49.9} & 52.9 & 41.1 & 65.6 & 27.0 & 53.2 & \multicolumn{1}{c|}{45.3} & 46.4 & \textcolor{red}{-6.5} & 3.46M\\ 


\multicolumn{1}{l|}{} & \multicolumn{1}{l|}{\cellcolor{g}BECoTTA+ (S)} &
\cellcolor{g}\textit{\textcolor{darkcerulean}{MoDE}} & 
\cellcolor{g}44.1 &
\cellcolor{g}69.5 &
\cellcolor{g}40.1 &
\cellcolor{g}56.8 &
\multicolumn{1}{c|}{\cellcolor{g}49.1} &
\cellcolor{g}51.9 &
\cellcolor{g}44.0 &
\cellcolor{g}69.4 &
\cellcolor{g}40.1 &
\cellcolor{g}56.9 &
\multicolumn{1}{c|}{\cellcolor{g}49.1} &
\cellcolor{g}51.9 &
\cellcolor{g}+0.0 & 
\cellcolor{g}0.12M \\ 

\multicolumn{1}{l|}{} & \multicolumn{1}{l|}{\cellcolor{gg}BECoTTA+ (M) } &
\cellcolor{gg}\textit{\textcolor{darkcerulean}{MoDE}} & 
\cellcolor{gg}45.6 & 
\cellcolor{gg}70.8 & 
\cellcolor{gg}42.6 & 
\cellcolor{gg}59.6 & 
\multicolumn{1}{c|}{\cellcolor{gg}\textbf{50.8}} & 
\cellcolor{gg}53.9 & 
\cellcolor{gg}45.6 & 
\cellcolor{gg}70.7 & 
\cellcolor{gg}42.5 & 
\cellcolor{gg}59.5 & 
\multicolumn{1}{c|}{\cellcolor{gg}\textbf{50.8}} & 
\cellcolor{gg}53.9 & 
\cellcolor{gg}+0.0 & 
\cellcolor{gg}0.77M \\ 

\multicolumn{1}{l|}{} & \multicolumn{1}{l|}{\cellcolor{ggg}BECoTTA+ (L) } &
\cellcolor{ggg}\textit{\textcolor{darkcerulean}{MoDE}} & 
\cellcolor{ggg}\textbf{45.7} &
\cellcolor{ggg}\textbf{71.4} &
\cellcolor{ggg}\textbf{43.7} &
\cellcolor{ggg}\textbf{59.6} &
\multicolumn{1}{c|}{\cellcolor{ggg}50.5} &
\cellcolor{ggg}\textbf{54.2} &
\cellcolor{ggg}\textbf{45.7} &
\cellcolor{ggg}\textbf{71.3} &
\cellcolor{ggg}\textbf{43.7} &
\cellcolor{ggg}\textbf{59.6} &
\multicolumn{1}{c|}{\cellcolor{ggg}50.6} &
\cellcolor{ggg}\textbf{54.2} &
\cellcolor{ggg}\textbf{+0.0} & 
\cellcolor{ggg}3.32M \\
\bottomrule
\end{tabular}%
}
\label{ref:main2_v2}
\end{table*}


%% file: tex/06_experiments.tex

\section{Experiments} 

We first introduce the datasets in~\Cref{sec:subsec:datasetup}, used for three continual segmentations, two classifications, and a domain generalization benchmark. 
Next, we describe the experimental setup in~\Cref{sec:subsec:expsetup}. 
Then, we provide our main results and analysis in~\Cref{sec:subsec:mainresult,sec:subsec:analysis}, respectively. More detailed results are in the Appendix.

\subsection{Datasets}\label{sec:subsec:datasetup} 

\textbf{Continual Disjoint Shifts (CDS) benchmark.}
To reflect various domain shifts, we adopt balanced weather shifts (CDS-\textit{Easy}) and imbalanced weather \& area shifts (CDS-\textit{Hard}) scenarios.
For the CDS-\textit{Easy}, we utilize the Cityscapes-ACDC setting used in previous work~\cite{cotta}: Cityscapes~\cite{cityscapes} is used as the source domain, and ACDC~\cite{acdc} as the target domain, consisting of four different weather types (fog, night, rain, snow).  
For the CDS-\textit{Hard}, we propose a new imbalanced scenario considering both weather and geographical domain shifts. We also add clear and overcast weather from BDD-100k~\cite{bdd} to the existing target domain to mimic the real-world variety. 

\textbf{Continual Gradual Shifts (CGS) benchmark.}
To construct gradually changing weather scenarios with blurry boundaries, we define sampling distributions per weather and perform uniform sampling. Next, we introduce four tasks containing blurred boundaries of weather, as illustrated in \Cref{fig:overall}. The detailed process is in Appendix (\Cref{append:data}). 


\textbf{Classification benchmark.}
We additionally evaluate classification scenarios on CIFAR10-CIFAR10C~\cite{cifar10} and CIFAR100-CIFAR100C~\cite{cifar100} with a \textit{non-ViT backbone}. 

\textbf{Domain Generalization (DG) benchmark.}
To demonstrate the versatility of BECoTTA, we conduct additional zero-shot experiments using the DG benchmark~\cite{choi2021robustnet}. 
This benchmark includes two large-scale real-world datasets (BDD-100k~\cite{bdd}, Mapillary~\cite{mapillary}) and two simulated datasets (GTAV~\cite{gtav}, Synthia~\cite{synthia}).

\input{table/cotta}
\subsection{Experimental Setting}\label{sec:subsec:expsetup} 

\textbf{Baselines.} We compare our model with strong continual test-time adaptation methods including TENT~\cite{tent}, CoTTA~\cite{cotta}, SAR~\cite{sar}, EcoTTA~\cite{ecotta}, VDP~\cite{vdp}, DePT~\cite{dept}, TTN~\cite{lim2023ttn}. More details are found in Appendix (\Cref{baseline}).

\textbf{Evaluation metric.}
All of the semantic segmentation results are reported mIoU in \%.  
For the overall scenarios, we repeat each task in 10 rounds (a few rounds are reported for visibility). Please refer to the Appendix (\Cref{append:qual}) for the whole results. 
For the classification tasks, we report the classification error rate (\%) following other baselines.

\input{table/blurry}

\textbf{Implementation details.} 
BECoTTA has a flexible architecture design, so it provides multiple variants according to the selection of the expert rank ($dim$), the location of MoDE, the number of experts ($N$),  and domain routers ($D$). 
Regarding $D$, we adopt D=1 for BECoTTA and D=4 for BECoTTA+. 
Regarding $dim$ and $N$, we categorize the results into three groups: S, M, and L. 
More specifically, we set four experts for S, and only inject MoDE into the last block of the encoder. 
Both M and L utilize six experts and inject MoDE into every block of the encoder. The difference between M and L is the $dim$ setting.
More variants are found in \Cref{table:hiddendim}.

For the CDS-\textit{Easy} scenario, we leverage the pre-trained Segformer-B5 as our source model, aligning with CoTTA~\cite{cotta}. For other scenarios, we opt for Segformer-B2.
Note that there is a difference between the two setups due to the size of Segformer, but we unify the Ours-S, M, and L architecture settings for all scenarios. 
To implement the initialization process, we warm up our architecture for 10 epochs like previous works~\cite{ecotta,lim2023ttn}. 
For the classification task, we adopt the non-ViT backbones, WideResNet-28 for CIFAR10-CIFAR10C and WideResNet-40 for CIFAR100-CIFAR100C, for a fair comparison with other baselines. We provide further implementation details in Appendix (\Cref{append:implement}). 

\textbf{Fairness with other baselines.} 
We report both \textit{w/o SDA} (i.e., BECoTTA) and \textit{w/ SDA} (i.e., BECoTTA+) results for all experiments.
In \textit{w/ SDA} setup, we perform a slight initialization while activating full model parameters for the baselines~\cite{cotta,tent,sar} that update full parameters or normalization layers only during CTTA. 
On the other hand, CTTA methods with parameter-efficient modules, such as EcoTTA and ours, perform initialization using SDA by updating these trainable modules only while freezing the pre-trained backbone weights. 

\input{table/cifar100}


\subsection{Main Results}\label{sec:subsec:mainresult}  

\textbf{\textit{CDS-\textit{Hard} (imbalanced weather \& area shifts)}.}
As shown in \Cref{ref:main2_v2} and \Cref{fig:main_performance}, all of our BECoTTA-S/M/L outperforms strong CTTA baselines with fewer parameters. 
In the case of \textit{w/o SDA}, although TENT and SAR utilize fewer parameters, they suffer from severe forgetting at 10 rounds. 
Otherwise, our BECoTTA achieves \textbf{+48.2\%}\textit{p}, \textbf{+7.7\%}\textit{p} improvement than TENT and SAR respectively at the last round. 
In addition, our BECoTTA(S) demonstrates 1.81\%\textit{p} gain using only \textbf{$\sim$98\%} fewer parameters (0.09M) than EcoTTA, while preserving the previous domain knowledge. 
Over CoTTA, all of our BECoTTA(S), (M), and (L) achieve \textbf{+1\%}\textit{p}, \textbf{+1.4\%}\textit{p} and \textbf{+2.1\%}\textit{p} increased performance using \textbf{608$\times$, 86$\times$} and \textbf{17$\times$} reduced parameter, respectively. 
In the case of \textit{w/ SDA}, our method surpasses other baselines that utilize \textit{\textcolor{crimson}{fully}} updated source models even only updating \textit{\textcolor{darkcerulean}{MoDE}} layers.
In particular, our BECoTTA+(L) shows a \textbf{16.8\%}\textit{p} improvement over \textit{w/ SDA} EcoTTA, which similarly updates only \textit{\textcolor{darkcerulean}{MetaNet}} as ours.

\textbf{\textit{CDS-\textit{Easy} (balanced weather shifts)}.}
As demonstrated in \Cref{ref:main1_v2}, BECoTTA achieves superior performance over other strong baselines. 
Compared within \textit{w/o SDA} only, our BECoTTA(S) outperforms EcoTTA (\textbf{+5\%}\textit{p}) and CoTTA (\textbf{+0.1\%}\textit{p}) by using only \textbf{2\%} and \textbf{0.1\%} number of parameters they used. 
Additionally, while BECoTTA(S) uses a similar level of parameters (0.09M) as TENT and SAR, we demonstrate a \textbf{+5\%}\textit{p} performance increase compared to them. 
Ultimately, our BECoTTA succeeds in achieving \textbf{+11.1\%}\textit{p} higher performance than the source-only.

\textbf{\textit{Continual Gradual Shifts (CGS)}.}
As shown in \Cref{ref:blurry}, we display the first round CGS scenario including four tasks. 
Even though the target domain is the same setting as CDS-\textit{Easy}, the overall performances are measured higher since the accessibility of previous domains. 
Our BECoTTA(L) achieves \textbf{+5.5\%}\textit{p} higher performance than EcoTTA with a similar number of parameters (3.16M). 
In this case, the input-dependent process of BECoTTA performs well in these blurry scenarios and ultimately shows \textbf{+13.4\%}\textit{p} improvement over the source model.

\textit{\textbf{Classification.}}
In addition to evaluating our method on segmentation tasks, we provide results of CIFAR100-to-CICAR100C classifications to further demonstrate the generalizability of BECoTTA. 
As shown in \Cref{table:cifar100cc}, BECoTTA (w/o SDA) consistently outperforms strong baselines, showing \textbf{-6.82\%}\textit{p} and \textbf{-2.47\%}\textit{p} reduction of the average error rate over CoTTA and EcoTTA, respectively. 
In particular, there are significant improvements over EcoTTA in \textit{Contrast} (\textbf{-8.57\%}\textit{p}) and \textit{Fog} (\textbf{-5.41\%}\textit{p}) which have similar attributes to the weather domain. 
See further results and analyses including computational efficiency in~\Cref{table:cifar10} and \Cref{table:cifar10effi}.

\textit{\textbf{Zero-shot Domain Generalization (DG).}}
We further demonstrate the versatility of our method through the zero-shot evaluation on four well-known driving datasets: BDD100k, Mapillary, CTAV, and Synthia. 
We compare the zero-shot performance of models with two different backbones, Deeplab v3+ and Segformer-B2. 
As shown in~\Cref{ref:dg}, our proposed method continuously outperforms strong baselines, demonstrating the competitive potential of the generalization ability over unseen domains.

\input{table/dg}

\subsection{Analyses and Ablations}\label{sec:subsec:analysis}

\begin{figure}[t]
    \centering
    \includegraphics[width=0.49\textwidth]{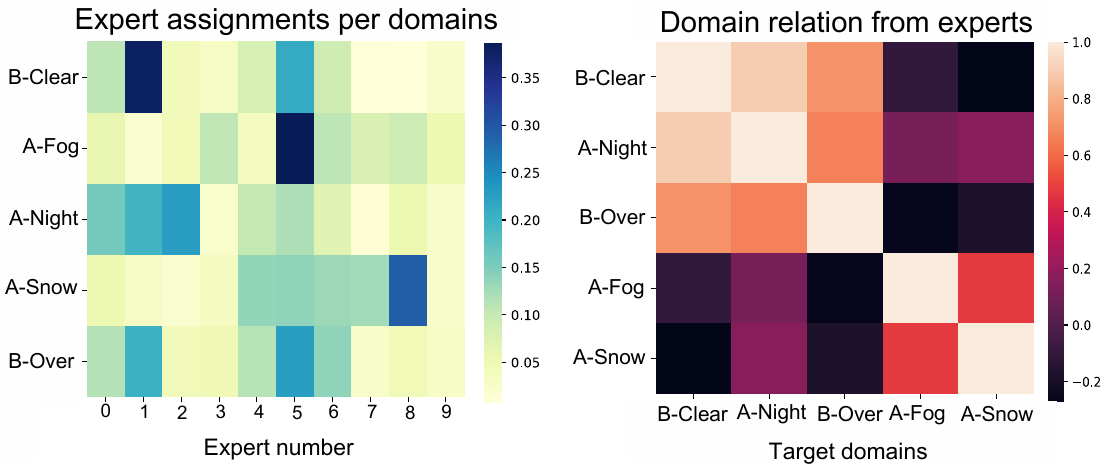}
    \caption{
    \textbf{Expert Analysis.} 
    \textbf{Left:} We visualize the frequency of ten expert selections for each domain during CTTA.
    Our frequency map shows co-selected and isolated experts in different domains. 
    \textbf{Right:} We interpret the similarity between target domains by visualizing the assignment weights from each domain-adaptive router.
    }
    \label{fig:qual_effect}
\vspace{-0.1in}
\end{figure}

\textbf{Experts analysis.} We represent an in-depth analysis of domain experts.
In ~\Cref{fig:qual_effect} (a), we visualize the frequency at which the domain expert is selected.
It is noteworthy that the weather scenes with similar visual contexts share similar experts. For instance, in the case of \textit{Clear} and \textit{Overcast} scene, domain experts \#1 and \#5 are commonly selected. 
Also, in the case of the \textit{night} scene, the distinct experts are selected compared to other scenes.
This faithfully represents our BECoTTA facilitates the cooperation and specialization among each domain expert. 
As illustrated in \Cref{fig:qual_effect} (b), we also derive the similarity between domains based on the selected experts. 
It is seen that \{\textit{Clear, Night, Overcast}\} and \{\textit{Fog, Snow}\} share visual context according to our ten domain experts.

\begin{figure}[t]
    \centering
    \begin{minipage}{0.48\linewidth}
        \centering
        \includegraphics[width=1.0\linewidth]{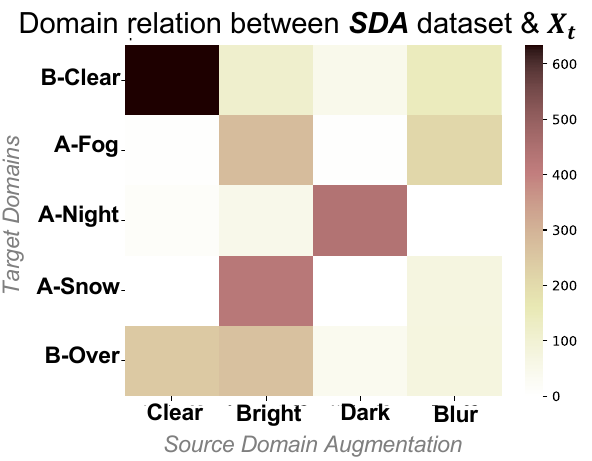}
        \vspace{-.5cm}
        \caption*{(a) Domain Relation}
    \end{minipage}
    \hspace{0.1cm}
    \begin{minipage}{0.48\linewidth}
        \centering
        \input{./table/sda_ablation}      
        \caption*{(b) Ablation of BECoTTA+}
    \end{minipage}
    
    \captionsetup{justification=justified} 
    \caption{\textbf{Domain Relation and Ablation study.} \textbf{(a)}: The domain relation between pre-defined SDA dataset and target domains.\textbf{(b)}: The ablation study for each element of BECoTTA+. AvgIoU is measured among 10 rounds.}
    \vspace{-0.15in}
    \label{fig:mix}
\end{figure}

\textbf{Ablation for each element.} 
We observe the variation of the main components: Domain Discriminator (DD), MoDE, and domain-expert synergy loss $\Theta(D;A)$. 
As shown in \Cref{fig:mix} right, relying solely on DD results in only \textbf{+0.04\%}\textit{p} improvement compared with the source only. 
However, when incorporating the MoDE and $\Theta(D;A)$, there are \textbf{+5.36\%}\textit{p} and \textbf{+6.12\%}\textit{p} improvement respectively, comparing with the source model. 

\begin{figure}[t]
    \centering
    \includegraphics[width=1.0\linewidth] {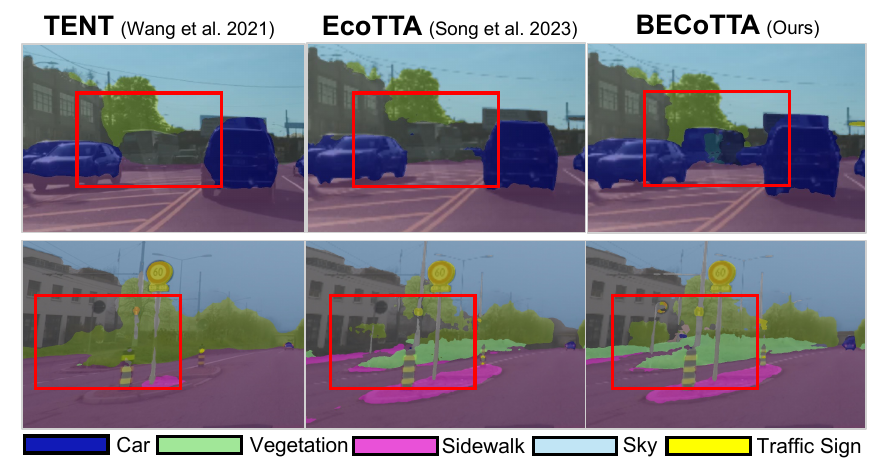}
    \caption{\textbf{Pseudo label Visualization.} Our BECoTTA generates more fine-grained and accurate labels than baselines.}
    \label{fig:fine}
\end{figure}

\textbf{Dependency on SDA.}
To validate the independence of BECoTTA on SDA, we perform an ablation with various combinations of augmented source domains. 
For example, D=2 (\textit{Source}, \textit{Night}) indicates SDA where the source domain and Night style augmented data (brightness adjusted to be darker), are included. 
As~\Cref{tab:d_ablation} represents, our MoDE layer allows a domain-specific adaptation regardless of the relevancy between SDA and target (test-time) domains. For instance, even if SDA only constitutes (\textit{Source}, \textit{Bright}) domains, BECoTTA+ responds to the \textit{\textbf{Night}} target domain effectively, resulting in \textbf{8.5\%}\textit{p} higher IoU in the night domain than w/o SDA.

\textbf{Relation between SDA and $\bm{X_t}$.} 
During the CTTA, we generate pseudo domain labels $d$ for each $X_t$ (target domains) through DD. 
These labels are used for initial domain-router assignment to facilitate domain-adaptive routing. 
Therefore, the relevance between the SDA and the target domain is crucial. 
In \Cref{fig:mix} left, we represent the relationship between pre-defined SDA and $X_t$, and our DD faithfully reflects this connection.

\textbf{Pseudo label analysis.} 
In ~\Cref{fig:fine}, we compare the generated pseudo labels after finishing the ten rounds.  
Our method exhibits robustness to forgetting in pseudo-label generation compared to other models. 
For other baselines, there is an erosion of minor labels by the effect of dominant labels (e.g., sky). 
However, our BECoTTA prevents such occurrences and effectively preserves pseudo labels as each round goes by, and it demonstrates significant efficacy in preserving fine-grained labels.

\begin{table}[t]
    \centering
    \caption{\textbf{Ablation study for the number of SDA domains.} }
    \renewcommand{\arraystretch}{1.5}
    \resizebox{\columnwidth}{!}{%
    \begin{tabular}{l|ccccc|c}
    \hline
        ~ & B-Clear & A-Fog & A-Night & A-Snow & B-Overcast & Avg \\ \hline
        CoTTA~\cite{cotta} & 43.3 & 67.3 & 34.8 & 56.9 & 48.8 & 50.2 \\
        EcoTTA~\cite{ecotta} & 41.9 & 66.1 & 31.5 & 55.3 & 46.2 & 48.2 \\ \hline
        BECoTTA  (w/o SDA) & 43.0 & 69.5 & 35.1 & 57.3 & 48.8 & 50.7 \\ 
        \textbf{D=2} (\textit{Source}, \textit{Night}) & 44.2 & 68.9 & 39.5 & 56.8 & 49.4 & 51.7 \\ 
        \textbf{D=2} (\textit{Source}, \textit{Bright}) & 43.9 & 69.2 & 38.1 & 57.7 & 49.3 & 51.6 \\ 
        \textbf{D=3} (\textit{Source}, \textit{Night}, \textit{Bright}) & 44.1 & 69.3 & 40.4 & 57.1 & 49.1 & 52.0 \\ \hline
    \end{tabular}
    }
    \label{tab:d_ablation}
\end{table}

%% file: table/cotta.tex
\begin{table*}[h]
\centering
\label{ref:main1_v2}
\renewcommand{\arraystretch}{1.3}

\setlength{\tabcolsep}{.6em}

\caption{\textbf{Results on \textit{CDS-Easy (balanced weather shifts)}.} We use the Cityscapes-to-ACDC benchmark, containing balanced weather shifts for target domains. For a fair comparison, we report both \textit{w/o WAD} and \textit{w/ WAD} performance of our method. The number of the parameters for DePT and VDP are not available as they do not provide the official codes.}

\resizebox{\linewidth}{!}{%

\begin{tabular}{lc|cccccccccccc|c|c}
\hline
\multicolumn{2}{l|}{Round} & \multicolumn{4}{c|}{1} & \multicolumn{4}{c|}{2} & \multicolumn{4}{c|}{3} & \multicolumn{2}{l}{} \\ \hline

\multicolumn{1}{l|}{Method} & \multicolumn{1}{c|}{Venue} &  Fog & Night & Rain & \multicolumn{1}{c|}{Snow} & Fog & Night & Rain & \multicolumn{1}{c|}{Snow} & Fog & Night & Rain & \multicolumn{1}{c|}{Snow} & \multicolumn{1}{c|}{Mean} & Parameter\\ \hline

\multicolumn{1}{l|}{Source only} & \multicolumn{1}{c|}{\textit{NIPS'21}} &  69.1 & 40.3 & 59.7 & \multicolumn{1}{c|}{57.8} & 69.1 & 40.3 & 59.7 & \multicolumn{1}{c|}{57.8} & 69.1 & 40.3 & 59.7 & \multicolumn{1}{c|}{57.8} & 56.7 & - \\

\multicolumn{1}{l|}{BN Stats Adapt~\cite{bnadapt}} & \multicolumn{1}{c|}{\textit{-}}  & 62.3 & 38.0 & 54.6 & \multicolumn{1}{c|}{53.0} & 62.3 & 38.0 & 54.6 & \multicolumn{1}{c|}{53.0} & 62.3 & 38.0 & 54.6 & \multicolumn{1}{c|}{53.0} & 52.0 & \textit{0.09M} \\ 

\multicolumn{1}{l|}{TENT~\cite{tent}} & \multicolumn{1}{c|}{\textit{ICLR'21}} & 69.0 & 40.2 & 60.1 & \multicolumn{1}{c|}{57.3} & 68.3 & 39.0 & 60.1 & \multicolumn{1}{c|}{56.3} & 67.5 & 37.8 & 59.6 & \multicolumn{1}{c|}{55.0} & 55.8  & \textit{0.09M} \\

\multicolumn{1}{l|}{CoTTA~\cite{cotta}} & \multicolumn{1}{c|}{\textit{CVPR'22}} & 70.9 & 41.2 & 62.4 & \multicolumn{1}{c|}{59.7} & 70.9 & 41.1 & 62.6 & \multicolumn{1}{c|}{59.7} & 70.9 & 41.0 & 62.7 & \multicolumn{1}{c|}{59.7} & 58.5 & \textbf{\textit{84.61M}} \\

\multicolumn{1}{l|}{SAR~\cite{sar}} & \multicolumn{1}{c|}{\textit{ICLR'23}} & 69.0 & 40.2 & 60.1 & \multicolumn{1}{c|}{57.3} & 69.0 & 40.3 & 60.0 & \multicolumn{1}{c|}{67.8} & 67.5 & 37.8 & 59.6 & \multicolumn{1}{c|}{55.0} & 55.8 & \textit{0.09M} \\

\multicolumn{1}{l|}{DePT~\cite{dept}} & \multicolumn{1}{c|}{\textit{ICLR'23}} &71.0 & 40.8 & 58.2 & \multicolumn{1}{c|}{56.8} & 68.2 & 40.0 & 55.4 & \multicolumn{1}{c|}{53.7} & 66.4 & 38.0 & 47.3 & \multicolumn{1}{c|}{47.2} & 53.5 & \textit{N/A} \\

\multicolumn{1}{l|}{VDP~\cite{vdp}} & \multicolumn{1}{c|}{\textit{AAAI'23}} &70.5 & 41.1 & 62.1 & \multicolumn{1}{c|}{59.5} & 70.4 & 41.1 & 62.2 & \multicolumn{1}{c|}{59.4} & 70.4 & 41.0 & 62.2 & \multicolumn{1}{c|}{59.4} & 58.2 & \textit{N/A} \\

\multicolumn{1}{l|}{EcoTTA~\cite{ecotta}} & \multicolumn{1}{c|}{\textit{CVPR'23}} &  68.5 & 35.8 & 62.1 & \multicolumn{1}{c|}{57.4} & 68.3 & 35.5 & 62.3 & \multicolumn{1}{c|}{57.4} & 68.1 & 35.3 & 62.3 & \multicolumn{1}{c|}{57.3} & 55.8 & \textit{3.46M} \\ 



\multicolumn{1}{l}{\cellcolor[HTML]{EEFFDD}{BECoTTA (S)}} & \multicolumn{1}{c|}{\cellcolor[HTML]{EEFFDD}{}} & \cellcolor[HTML]{EEFFDD}{71.3} & \cellcolor[HTML]{EEFFDD}{41.1} & \cellcolor[HTML]{EEFFDD}{62.4} & \multicolumn{1}{c|}{\cellcolor[HTML]{EEFFDD}{59.8}} &\cellcolor[HTML]{EEFFDD}{71.3}  & \cellcolor[HTML]{EEFFDD}{41.1} & \cellcolor[HTML]{EEFFDD}{62.4} & \multicolumn{1}{c|}{\cellcolor[HTML]{EEFFDD}{59.8}} & \cellcolor[HTML]{EEFFDD}{71.4} & \cellcolor[HTML]{EEFFDD}{41.1} & \cellcolor[HTML]{EEFFDD}{62.4} & \multicolumn{1}{c|}{\cellcolor[HTML]{EEFFDD}{59.8}} & \cellcolor[HTML]{EEFFDD}{58.6} & \cellcolor[HTML]{EEFFDD}{\textit{0.09M}} \\
\multicolumn{1}{l}{\cellcolor[HTML]{EEFFDD}{~~~~~~~~~~~~~~~~~~~~~~~~~~~~~~~~~~~\textit{+ SDA}}} & \multicolumn{1}{c|}{\cellcolor[HTML]{EEFFDD}{}} & \cellcolor[HTML]{EEFFDD}{72.0} & \cellcolor[HTML]{EEFFDD}{45.4} & \cellcolor[HTML]{EEFFDD}{63.7} & \multicolumn{1}{c|}{\cellcolor[HTML]{EEFFDD}{60.0}} & \cellcolor[HTML]{EEFFDD}{71.7} & \cellcolor[HTML]{EEFFDD}{45.2} & \cellcolor[HTML]{EEFFDD}{63.6} & \multicolumn{1}{c|}{\cellcolor[HTML]{EEFFDD}{60.1}} & \cellcolor[HTML]{EEFFDD}{71.7} &\cellcolor[HTML]{EEFFDD}{45.4}  & \cellcolor[HTML]{EEFFDD}{63.6} &\multicolumn{1}{c|}{\cellcolor[HTML]{EEFFDD}{60.1}}  & \multicolumn{1}{c|}{\cellcolor[HTML]{EEFFDD}{60.2}} & \cellcolor[HTML]{EEFFDD}{\textit{0.12M}} \\

\multicolumn{1}{l}{\cellcolor[HTML]{DDEECC}{BECoTTA (M)}} & \multicolumn{1}{c|}{\cellcolor[HTML]{DDEECC}{}} & \cellcolor[HTML]{DDEECC}{72.3} & \cellcolor[HTML]{DDEECC}{42.0} & \cellcolor[HTML]{DDEECC}{63.5} & \multicolumn{1}{c|}{\cellcolor[HTML]{DDEECC}{60.1}} & \cellcolor[HTML]{DDEECC}{72.4} & \cellcolor[HTML]{DDEECC}{41.9} & \cellcolor[HTML]{DDEECC}{63.5} & \multicolumn{1}{c|}{\cellcolor[HTML]{DDEECC}{60.2}} & \cellcolor[HTML]{DDEECC}{72.3} & \cellcolor[HTML]{DDEECC}{41.9} & \cellcolor[HTML]{DDEECC}{63.6} & \multicolumn{1}{c|}{\cellcolor[HTML]{DDEECC}{60.2}} & \cellcolor[HTML]{DDEECC}{59.5} & \cellcolor[HTML]{DDEECC}{\textit{2.15M}} \\
\multicolumn{1}{l}{\cellcolor[HTML]{DDEECC}{~~~~~~~~~~~~~~~~~~~~~~~~~~~~~~~~~~~\textit{+ SDA}}} & \multicolumn{1}{c|}{\cellcolor[HTML]{DDEECC}{}} &  \cellcolor[HTML]{DDEECC}{71.8} & \cellcolor[HTML]{DDEECC}{48.0} & \cellcolor[HTML]{DDEECC}{66.3} & \multicolumn{1}{c|}{\cellcolor[HTML]{DDEECC}{62.0}} & \cellcolor[HTML]{DDEECC}{71.7} & \cellcolor[HTML]{DDEECC}{47.7} & \cellcolor[HTML]{DDEECC}{66.3} & \multicolumn{1}{c|}{\cellcolor[HTML]{DDEECC}{61.7}} & \cellcolor[HTML]{DDEECC}{71.8} & \cellcolor[HTML]{DDEECC}{47.7} & \cellcolor[HTML]{DDEECC}{66.3} & \multicolumn{1}{c|}{\cellcolor[HTML]{DDEECC}{61.9}} & \cellcolor[HTML]{DDEECC}{61.9} & \cellcolor[HTML]{DDEECC}{\textit{2.70M}} \\

\multicolumn{1}{l}{\cellcolor[HTML]{CCDDBB}{BECoTTA (L)}} & \multicolumn{1}{c|}{\cellcolor[HTML]{CCDDBB}{}} & \cellcolor[HTML]{CCDDBB}{71.5} & \cellcolor[HTML]{CCDDBB}{42.6} & \cellcolor[HTML]{CCDDBB}{63.2} & \multicolumn{1}{c|}{\cellcolor[HTML]{CCDDBB}{59.1}} & \cellcolor[HTML]{CCDDBB}{71.5} & \cellcolor[HTML]{CCDDBB}{42.6} & \cellcolor[HTML]{CCDDBB}{63.2} & \multicolumn{1}{c|}{\cellcolor[HTML]{CCDDBB}{59.1}} & \cellcolor[HTML]{CCDDBB}{71.5} & \cellcolor[HTML]{CCDDBB}{42.5} & \cellcolor[HTML]{CCDDBB}{63.2} & \multicolumn{1}{c|}{\cellcolor[HTML]{CCDDBB}{59.1}} & \cellcolor[HTML]{CCDDBB}{59.1} & \cellcolor[HTML]{CCDDBB}{\textit{11.31M}} \\
\multicolumn{1}{l}{\cellcolor[HTML]{CCDDBB}{~~~~~~~~~~~~~~~~~~~~~~~~~~~~~~~~~~~\textit{+ SDA}}} & \multicolumn{1}{c|}{\cellcolor[HTML]{CCDDBB}{}} & \cellcolor[HTML]{CCDDBB}{\textbf{72.7}} & \cellcolor[HTML]{CCDDBB}{\textbf{49.5}} & \cellcolor[HTML]{CCDDBB}{\textbf{66.3}} & \multicolumn{1}{c|}{\cellcolor[HTML]{CCDDBB}{\textbf{63.1}}} & \cellcolor[HTML]{CCDDBB}{\textbf{72.6}} & \cellcolor[HTML]{CCDDBB}{\textbf{49.4}} & \cellcolor[HTML]{CCDDBB}{\textbf{66.3}} & \multicolumn{1}{c|}{\cellcolor[HTML]{CCDDBB}{\textbf{62.8}}} & \cellcolor[HTML]{CCDDBB}{\textbf{72.5}} & \cellcolor[HTML]{CCDDBB}{\textbf{49.7}} & \cellcolor[HTML]{CCDDBB}{\textbf{66.2}} & \multicolumn{1}{c|}{\cellcolor[HTML]{CCDDBB}{\textbf{63.1}}} & \cellcolor[HTML]{CCDDBB}{\textbf{63.0}} & \cellcolor[HTML]{CCDDBB}{\textit{11.86M}}\\ \hline

\end{tabular}
}%

\end{table*}

%% file: table/blurry.tex
\begin{table}[t!]
\centering
\renewcommand{\arraystretch}{1.4}
\caption{\textbf{Results on \textit{Continual Gradual Shifts (CGS)}.} We construct the novel gradual shifts scenario using CDS-Easy target domains. 
}
\vspace{-0.05in}
\resizebox{0.95\linewidth}{!}{%
\begin{tabular}{l|cccc|c|c}
\hline
 & Task 1 & Task 2 & Task 3 & Task 4 & Mean & Parameter\\ \hline \hline
 
Source &  57.93 & 44.15 & 55.54 & 54.73 & 53.09 & - \\

TENT~\cite{tent} &  58.12 & 44.67 & 56.35 & 55.26 & 53.60 & \textit{0.02M}\\
SAR~\cite{sar}  & 57.95 & 44.23 & 55.67 & 54.92 & 53.19 &\textit{0.02M}\\
EcoTTA~\cite{ecotta} & 62.15 & 47.60 & 59.70 & 58.70 & 57.04 & \textit{3.46M}\\  

\rowcolor[HTML]{EEFFDD} 
BECoTTA (S)  &61.85  &46.95   &57.64   &56.96   & 55.85 & \textit{0.09M} \\

\rowcolor[HTML]{EEFFDD} 
~~~~~~~~~~~~~~~~~+ \textit{SDA} & 62.09  &51.08  &59.90   &57.72   &57.69 & \textit{0.12M}   \\   

\rowcolor[HTML]{DDEECC} 
BECoTTA (M)   & 60.49 & 46.20 & 58.24 & 57.45 & 55.60 & \textit{0.63M} \\
\rowcolor[HTML]{DDEECC} 
~~~~~~~~~~~~~~~~~+ \textit{SDA} & 64.04 & 53.25 & 60.66 & 58.55 & 59.13 & \textit{0.77M} \\  

\rowcolor[HTML]{CCDDBB} 
BECoTTA (L)   & 62.55  & 47.72   & 59.30  &  59.02  &  57.13  & \textit{3.16M}\\

\rowcolor[HTML]{CCDDBB} 
~~~~~~~~~~~~~~~~~+ \textit{SDA}  & \textbf{64.62}  & \textbf{53.54} & \textbf{62.59}  & \textbf{60.17}  & \textbf{60.23} & \textit{3.31M} \\  \hline

\end{tabular}%
} 
\vspace{-0.15in}     
\label{ref:blurry} 
\end{table}


%% file: table/cifar100.tex
\begin{table*}[!ht]
    \centering
    \renewcommand{\arraystretch}{1.3}
    \setlength{\tabcolsep}{.33em}
    \caption{\textbf{Classification error rate (\%) for CIFAR100-to-CIFAR100C with severity level 5}. Results are evaluated on WideResNet-40. 
    }
    {\scriptsize
    \begin{tabular}{l|ccccccccccccccc|c}
    \hline
        Method & Gaus. & Shot & Impu. & Defo. & Glas. & Moti. & Zoom & Snow & Fros. & Fog & Brig. & Cont. & Elas. & Pixe. & Jpeg & Avg. err \\ \hline
        Source & 80.1 & 77.0 & 76.4 & 59.9 & 77.6 & 64.2 & 59.3 & 64.8 & 71.3 & 78.3 & 48.1 & 83.4 & 65.8 & 80.4 & 59.2 & 69.7 \\ 
        tBN~\cite{tbn} & 45.9 & 45.6 & 48.2 & 33.6 & 47.9 & 34.5 & 34.1 & 40.3 & 40.4 & 47.1 & 31.7 & 39.7 & 42.7 & 39.2 & 45.6 & 41.1 \\ 
        Single do. TENT~\cite{tent} & 41.2 & 40.6 & 42.2 & 30.9 & 43.4 & 31.8 & 30.6 & 35.3 & 36.2 & 40.1 & 28.5 & 35.5 & 39.1 & 33.9 & 41.7 & 36.7 \\ 
        Continual TENT~\cite{tent} & 41.2 & 38.2 & 41.0 & 32.9 & 43.9 & 34.9 & 33.2 & 37.7 & 37.2 & 41.5 & 33.2 & 37.2 & 41.1 & 35.9 & 45.1 & 38.3 \\ 
        TTT++~\cite{ttt} & 46.0 & 45.4 & 48.2 & 33.5 & 47.7 & 34.4 & 33.8 & 39.9 & 40.2 & 47.1 & 31.8 & 39.7 & 42.5 & 38.9 & 45.5 & 41.0 \\ 
        SWRNSP~\cite{swr} & 42.4 & 40.9 & 42.7 & 30.6 & 43.9 & 31.7 & 31.3 & 36.1 & 36.2 & 41.5 & 28.7 & 34.1 & 39.2 & 33.6 & 41.3 & 36.6 \\ 
        NOTE~\cite{note} & 50.9 & 47.4 & 49.0 & 37.3 & 49.6 & 37.3 & 37.0 & 41.3 & 39.9 & 47.0 & 35.2 & 34.7 & 45.2 & 40.9 & 49.9 & 42.8 \\ 
        EATA~\cite{eata} & 41.6 & 39.9 & 41.2 & 31.7 & 44.0 & 32.4 & 31.9 & 36.2 & 36.8 & 39.7 & 29.1 & 34.4 & 39.9 & 34.2 & 42.2 & 37.1 \\ 
        CoTTA~\cite{cotta} & 43.5 & 41.7 & 43.7 & 32.2 & 43.7 & 32.8 & 32.2 & 38.5 & 37.6 & 45.9 & 29.0 & 38.1 & 39.2 & 33.8 & 39.4 & 38.1 \\ 
        EcoTTA~\cite{ecotta} & 42.7 & 39.6 & 42.4 & 31.4 & 42.9 & 31.9 & 30.8 & 35.1 & 34.8 & 40.7 & 28.1 & 35.0 & 37.5 & 32.1 & 40.5 & 36.4 \\ 
        \cellcolor[HTML]{DDEECC}BECoTTA (w/o SDA) & \cellcolor[HTML]{DDEECC}42.1 & \cellcolor[HTML]{DDEECC}\textbf{38.0} & \cellcolor[HTML]{DDEECC}42.2 & \cellcolor[HTML]{DDEECC}\textbf{30.2} & \cellcolor[HTML]{DDEECC}\textbf{42.9} & \cellcolor[HTML]{DDEECC}\textbf{31.7} & \cellcolor[HTML]{DDEECC}\textbf{29.8} & \cellcolor[HTML]{DDEECC}\textbf{35.1} & \cellcolor[HTML]{DDEECC}\textbf{33.9} & \cellcolor[HTML]{DDEECC}\textbf{38.5} & \cellcolor[HTML]{DDEECC}\textbf{27.9} & \cellcolor[HTML]{DDEECC}\textbf{32.0} & \cellcolor[HTML]{DDEECC}\textbf{36.7} & \cellcolor[HTML]{DDEECC}\textbf{31.6} & \cellcolor[HTML]{DDEECC}39.9 & \cellcolor[HTML]{DDEECC}\textbf{35.5} \\ \hline
    \end{tabular}
    }
    \vspace{-.2cm}
\label{table:cifar100cc}
\end{table*}


%% file: table/dg.tex
\begin{table}[t!]
\centering
\renewcommand{\arraystretch}{1.5}
\caption{\textbf{Results on Zero-shot Domain Generalization.} We compare the zero-shot performance of our method with strong TTA methods on four unseen domains. 
}
\resizebox{\linewidth}{!}{%
{\normalsize
\begin{tabular}{c|l|c|c|c|c|c}
\hline
\rowcolor[HTML]{FFFFFF} 
Source model & \multicolumn{1}{c|}{Method} & \textbf{BDD100k} & \textbf{Mapillary} & \textbf{GTAV} & \textbf{Synthia} & \textbf{Avg} \\ \hline\hline
 & Source & 43.50 & 54.37 & 43.71 & 22.78 & 41.09 \\
 & BN Adapt~\cite{bnadapt} & 43.60 & 47.66 & 43.22 & 25.72 & 40.05 \\
 & TBN & 43.12 & 47.61 & 42.51 & 25.71 & 39.74 \\
 & TENT~\cite{tent} & 43.30 & 47.80 & 43.57 & 25.92 & 40.15 \\
 & SWR~\cite{swr} & 43.40 & 47.95 & 42.88 & 25.97 & 40.05 \\
\multirow{-6}{*}{Deeplab v3+} & TTN~\cite{lim2023ttn} & 48.85 & 59.09 & 46.71 & 29.16 & 45.95 \\ \hline

 & Source & 47.33 & 58.59 & 49.65 & 27.59 & 45.79 \\
 & TENT~\cite{tent} & 46.23 & 58.13 & 49.69 & 27.53 & 45.40 \\
 & SAR~\cite{sar} & 47.41 & 58.59 & 49.73 & 27.63 & 45.84 \\
 
 & \cellcolor[HTML]{DDEECC}BECoTTA (M) & \cellcolor[HTML]{DDEECC}\textbf{50.79} & \cellcolor[HTML]{DDEECC}\textbf{61.48} & \cellcolor[HTML]{DDEECC}\textbf{52.42} & \cellcolor[HTML]{DDEECC}\textbf{29.27} & \cellcolor[HTML]{DDEECC}\textbf{48.49} \\
 
\multirow{-5}{*}{Segformer-B2} & \cellcolor[HTML]{DDEECC}~~~~~~~~~~~~~~~~\textit{+ SDA} & \cellcolor[HTML]{DDEECC}\textbf{52.37} & \cellcolor[HTML]{DDEECC}\textbf{61.84} & \cellcolor[HTML]{DDEECC}\textbf{52.62} & \cellcolor[HTML]{DDEECC}\textbf{29.65} & \cellcolor[HTML]{DDEECC}\textbf{49.12} \\ 

 

\hline
\end{tabular}%
}
\label{ref:dg}
}
\vspace{-0.14in}
\end{table}

%% file: table/sda_ablation.tex
\renewcommand{\arraystretch}{1.2}
\resizebox{1.0\textwidth}{!}{%
{\footnotesize
\begin{tabular}{ccc|c}
\hline
DD & MoDE & $\Theta(D;A)$ & Avg IoU \\ \hline
&  &  & 51.14 \\
\ding{52} &  &  & 51.18 \\
\ding{52} & \ding{52} &  & 53.87 \\
\ding{52} & \ding{52} & \ding{52} & \textbf{54.27} \\
\hline
\end{tabular}%
}%
}%
\vspace{-0.1in}



%% file: tex/07_conclusion.tex
\section{Conclusion}

We propose BECoTTA, an efficient yet powerful approach for CTTA, mainly consisting of Mixture-of-Domain Low-rank Experts (MoDE). Our MoDE has two key components: (i) domain-adaptive routing, and (ii) domain-expert synergy loss to maximize the dependency between each domain and expert. We show that our BECoTTA outperforms other SoTA continual TTA models and exhibits significant efficiency with fewer parameters and memory. Besides, ours shows strong potential for zero-shot domain generalization tasks. To facilitate the understanding of our proposed method, we extensively provide various analyses, including ablations of each component of BECoTTA and WAD strategies, and visualize the obtained pseudo labels and the relationships between domains and experts.


\section*{Impact Statement}
In this work, we suggest BECoTTA and verify the superiority of performance and effectiveness. 
Due to its efficiency, our BECoTTA is highly effective when deployed on real-world embodied devices. 
This is particularly true in autonomous driving environments, where efficient adaptation is crucial. 
Moreover, it is freely applied to various real-world application branches, including health care and the medical field, which require continual adaptation.  
Therefore, we are confident that BECoTTA will have a significant impact on the practical application. 
We hope that our research focusing on efficiency contributes to the field of CTTA.

\section*{Acknowledgements}
This work was supported by Institute for Information \& communications Technology Promotion (IITP) grant funded by the Korea government (MSIP) (No.2019-0-00075 Artificial Intelligence Graduate School Program (KAIST), No.2022-0-00713), the National Research Foundation of Korea (NRF) grant funded by the Korea government (MSIT)(No. RS-2023-00256259).

%% file: tex/Appendix.tex
In this Appendix, we present the detailed material for a better understanding:  

First, we provide additional information about CTTA baselines~\Cref{baseline} and implementation details~\Cref{append:implement}.
The additional experiment results in up to 10 rounds, including \textit{CDS-Easy}, \textit{CDS-Hard}, CGS scenarios are provided at ~\Cref{append:qual}. 
Moreover, we evaluate the adaptation performance of BECoTTA in \textit{CIFAR10-CIFAR10C} and \textit{CIFAR100-CIFAR100C} classification scenarios. 
We also provide various ablation studies with diverse combinations of our architectures. 
At the end, more detail about the data construction process is provided in ~\Cref{append:data}.

\section{Baselines}\label{baseline}     
In this section, we provide the details of the TTA baselines we use in our main paper. We illustrate the details of other CTTA baselines in~\Cref{fig:comparison}. 

\textbf{CoTTA}~\cite{cotta} is a landmark work that proposed weight, augmentation averaged predictions, and stochastic restoration based on the mean-teacher framework. 
We utilize the official codes based on \texttt{mmsegmentation} that CoTTA author provided.\footnote{\url{https://github.com/qinenergy/cotta/issues/6}} 

\textbf{TENT}~\cite{tent} stands out as the pioneering approach to entropy minimization during testing, aiming to adapt to data shifts without the need for additional losses or data.
We follow the above implementation from CoTTA authors.

\textbf{SAR}~\cite{sar} point out a sharpness-aware entropy minimization that mitigates the impact of specific noisy test samples characterized by substantial gradients.
As SAR has not been specifically validated in segmentation scenarios, we refer to their code base \footnote{\url{https://github.com/mr-eggplant/SAR}} and reimplement it in the \texttt{mmsegmentation} framework.

\textbf{EcoTTA}~\cite{ecotta} propose memory-efficient architecture using meta networks. 
We believe there are similarities between our model and EcoTTA, particularly in emphasizing efficiency through the activation of small parts of the source model.
To ensure a fair comparison, we align EcoTTA's source model with our ViT-based Segformer. 
Behind each stage of Segformer, we insert their meta networks only four times in the source model ($K$=4). 



\section{Experiment Details}\label{append:implement}

\subsection{Implementation Details}
We provide the implementation details utilized in our experiments in~\Cref{table:ours_params,table:baseline_params}.

\begin{table}[h]
        \centering
        \caption{\textbf{Baseline method hyperparameters.}}
        \renewcommand{\arraystretch}{1.2}
        {\small
        \begin{tabular}{l|l}
        \hline
        EcoTTA~\cite{ecotta} & $K$=4, $\lambda$=0.5, $H_0$=0.4 \\
        Ours & $\lambda_d$=0.1, $\lambda_m$=0.0005, $\kappa$=0.4 \\ \hline
        \end{tabular}%
        }%
       \label{table:baseline_params}
\end{table}


\begin{table}[h]
        \centering
        \caption{\textbf{Our method hyperparameters.}}
        \renewcommand{\arraystretch}{1.2}
        {\footnotesize
            \begin{tabular}{l|cc}
            \hline
            \multicolumn{1}{c|}{} & Warm-up & TTA \\ \hline
            Dataset & SDA & Target domains \\
            Optimizer & AdamW & Adam \\
            Optimizer momentum & \multicolumn{2}{c}{\small{$(\beta_1, \beta_2)=(0.9, 0.999)$}} \\
            Epoch & 10 & Online \\
            Batch size & \multicolumn{2}{c}{1} \\
            Learning rate & 0.00006 & 0.00006/100 \\
            Label accessibility & Yes & No \\ \hline
            \end{tabular}%
        }%
        \label{table:ours_params}
     
\end{table}

\subsection{The details of BECoTTA}
\textbf{The size of BECoTTA.} Our BECoTTA has a flexible architecture design, which means we adjust the number of TTA parameters freely in various ways, depending on factors such as rank $r$, the insertion position of MoDE, and the number of experts $N$ in each MoDE.
(All of the `parameters' in our main paper mean the number of updated parameters in the TTA process.)
For the main experiments, we adopt four experts for S, and only injected MoDE into the last block. 
Both M and L utilize six experts for MoDE, with the only difference being the rank.

\textbf{Details of w/ \& w/o SDA.}
To ensure a fair comparison, we conduct all experiments in w/ \& w/o SDA settings. 
In the case w/o SDA, it is impossible to collect priors for domain candidates, therefore we adopt a single router $G$, similar to the conventional stochastic routing~\cite{thor}. 
In settings w/ SDA, as described in the paper, we employ domain-adaptive routing and a domain-experts synergy loss. This approach maximizes the effectiveness of BECoTTA.

\subsection{Algorithm of BECoTTA}
To clarify our whole CTTA process, we provide the whole pipeline of BECoTTA at~\Cref{alg:becotta_algo}. 
According to the different initialization steps (Line 1~9), BECoTTA can be initialized in various ways including SDA or not.

\begin{figure*}[h]
\centering
\begin{minipage}{0.65\textwidth}

\begin{algorithm}[H]
\caption{Continual Test-time Adaptation Pipeline}
\label{alg:becotta_algo}
\textbf{Input:} Source domain $X_s$, a sequence of target domains $X_t = \{X_t^1, X_t^2, \ldots\}$, source model $f$, trainable parts of MoDE $W_g^d$, $W_{\text{noise}}^d$, $W^{\text{down}}$, $W^{\text{up}}$, number of experts $N$, number of domain routers $D$.
\vspace{0.5ex} 
\begin{algorithmic}[1]
\small 
\STATE \texttt{\# Initialization}
\IF{SDA init}
    \STATE SDA = DomainAugment($X_s$)
    \STATE Update $W_g^d$, $W_{\text{noise}}^d$, $W^{\text{down}}$, $W^{\text{up}}$, Domain Discriminator $DD$ using SDA with $L_{\text{init}}$ 
\ELSIF{Source init}
    \STATE Update $W_g^d$, $W_{\text{noise}}^d$, $W^{\text{down}}$, $W^{\text{up}}$ using Source domain $X_s$ with $L_{\text{seg}}$
\ELSE
    \STATE Randomly initialize $W_g^d$, $W_{\text{noise}}^d$, $W^{\text{down}}$, $W^{\text{up}}$
\ENDIF

\vspace{0.5ex} 

\STATE \texttt{\# CTTA}
\FOR{target domain index $c = 1, 2, \ldots$}
    \FOR{minibatch $x \sim X_t^c$}
        \IF{SDA init}
            \STATE $d = DD(x)$
        \ELSE
            \STATE $d = \text{Uniform}(0, D)$
        \ENDIF
        \FOR{ViT block $\mathbb{B} \sim f$}
            \STATE $x = \mathbb{B}(x)$
            \STATE $h_d(x) \leftarrow G_d(x), W^{\text{down}}, W^{\text{up}}$ 
            \STATE $x \leftarrow x + h_d(x)$
        \ENDFOR
        \STATE Update $f$ with $L_{\text{tta}}$ 
    \ENDFOR
\ENDFOR
\end{algorithmic}
\end{algorithm}

\end{minipage}
\end{figure*}

\section{Additional Results}\label{append:qual}

\subsection{Results up to round 10}
\begin{figure*}[h]
    \centering
    \caption{\textbf{Pseudo labels from finished training up to round 10 in the CDS-Hard (imbalanced weather \& area shifts) scenario.} We visualize pseudo labels from BDD100k and ACDC datasets with other baselines. Our BECoTTA generates more fine-grained labels than other baselines.}
    {\includegraphics[width=.95\linewidth] {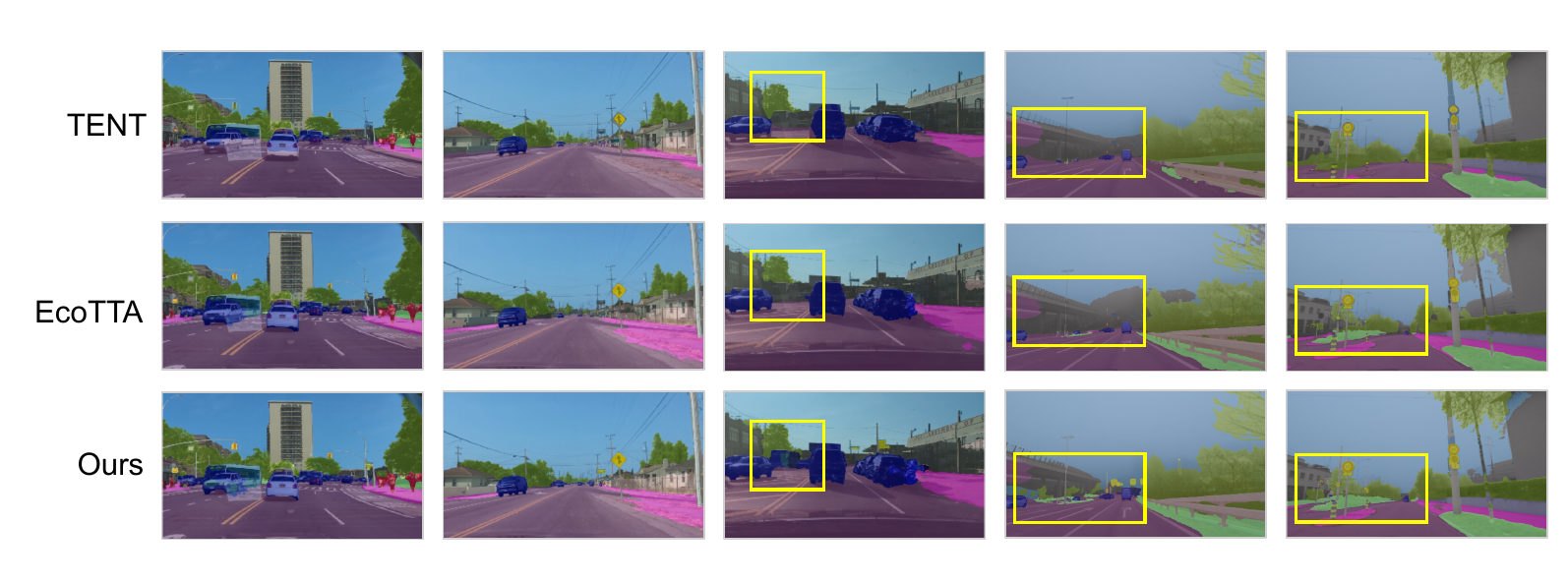}}       
    \label{fig:pseudos_appendix}
    \vspace{-0.7cm}
\end{figure*}

Following the previous works, such as CoTTA~\cite{cotta}, we repeat ten rounds to simulate long-term continual domain shifts. 
Therefore, as shown in \Cref{table:main1_ten}, \Cref{table:main2_ten}, and \Cref{table:blurry_ten}, we provide the whole performance up to 10 rounds for each CDS-\textit{Easy}, CDS-\textit{Hard}, and \textit{Continual Gradual Shifts (CGS)} scenarios. 
We also illustrate more qualitative results in ~\Cref{fig:pseudos_appendix}. 




\begin{table*}[t]
\centering
\renewcommand{\arraystretch}{1.3}
\setlength{\tabcolsep}{.23em}
\caption{\textbf{Quantitative results of \textit{CDS-Easy (balanced weather shifts)}.} We conduct experiments with Cityscapes-to-ACDC benchmarks which contain the weather shifts in the target domains. For a fair comparison, we report both \textit{w/o SDA} and \textit{w/ SDA} performance of our models. }

\resizebox{0.95\linewidth}{!}{%
\begin{tabular}{lc|cccccccccccccccc|c}
\hline
\multicolumn{2}{l|}{\textbf{Round}} & \multicolumn{4}{c|}{\textbf{1}} & \multicolumn{4}{c|}{\textbf{3}} & \multicolumn{4}{c|}{\textbf{7}} &  \multicolumn{4}{c|}{\textbf{10}} & \multicolumn{1}{l}{} \\ \hline

\multicolumn{1}{l|}{Method} & Venue & Fog & Night & Rain & \multicolumn{1}{c|}{Snow} 
& Fog & Night & Rain & \multicolumn{1}{c|}{Snow} 
& Fog & Night & Rain & \multicolumn{1}{c|}{Snow} 
& Fog & Night & Rain & Snow & Mean \\ \hline

\multicolumn{1}{l|}{Source only} & \textit{NIPS'21} & 69.1 & 40.3 & 59.7 & \multicolumn{1}{c|}{57.8} 
& 69.1 & 40.3 & 59.7 & \multicolumn{1}{c|}{57.8} 
& 69.1 & 40.3 & 59.7 & \multicolumn{1}{c|}{57.8} 
& 69.1 & 40.3 & 59.7 & 57.8 & 56.7 \\

\multicolumn{1}{l|}{BN Stats Adapt~\cite{bnadapt}} & \textit{-} & 62.3 & 38.0 & 54.6 & \multicolumn{1}{c|}{53.0} 
& 62.3 & 38.0 & 54.6 & \multicolumn{1}{c|}{53.0} 
& 62.3 & 38.0 & 54.6 & \multicolumn{1}{c|}{53.0} 
& 62.3 & 38.0 & 54.6 & 53.0 & 52.0 \\ 

\multicolumn{1}{l|}{Continual TENT~\cite{tent}} & \textit{ICLR'21} & 69.0 & 40.2 & 60.1 & \multicolumn{1}{c|}{57.3} 
& 68.3 & 39.0 & 60.1 & \multicolumn{1}{c|}{56.3} 
& 64.2 & 32.8 & 55.3 & \multicolumn{1}{c|}{50.9}     
& 61.8 & 29.8 & 51.9 & 47.8 & 52.3 \\                

\multicolumn{1}{l|}{CoTTA~\cite{cotta}} & \textit{CVPR'22} 
& 70.9 & 41.2 & 62.4 & \multicolumn{1}{c|}{59.7} 
& 70.9 & 41.0 & 62.7 & \multicolumn{1}{c|}{59.7}      
& 70.9 & 41.0 & 62.8 & \multicolumn{1}{c|}{59.7}    
& 70.8 & 41.0 & 62.8 & 59.7 & 58.6 \\

\multicolumn{1}{l|}{SAR~\cite{sar}} & \textit{ICLR'23} 
& 69.0 & 40.2 & 60.1 & \multicolumn{1}{c|}{57.3} 
& 69.1 & 40.3 & 60.0 & \multicolumn{1}{c|}{57.8} 
& 69.1 & 40.2 & 60.3 & \multicolumn{1}{c|}{57.9} 
& 69.1 & 40.1 & 60.5 & 57.9 & 56.8 \\

\multicolumn{1}{l|}{EcoTTA~\cite{ecotta}} & \textit{CVPR'23} 
& 68.5 & 35.8 & 62.1 & \multicolumn{1}{c|}{57.4} 
& 68.1 & 35.3 & 62.3 & \multicolumn{1}{c|}{57.3} 
& 67.2 & 34.2 & 62.0 & \multicolumn{1}{c|}{56.9} 
& 66.4 & 33.2 & 61.3 & 56.3 & 55.2 \\ 

\multicolumn{1}{l}{\cellcolor[HTML]{EEFFDD}{BECoTTA (Ours)-S}} & \cellcolor[HTML]{EEFFDD}{}
& \cellcolor[HTML]{EEFFDD}{71.3} & \cellcolor[HTML]{EEFFDD}{41.1} & \cellcolor[HTML]{EEFFDD}{62.4} & \multicolumn{1}{c|}{\cellcolor[HTML]{EEFFDD}{59.8}}
& \cellcolor[HTML]{EEFFDD}{71.4} & \cellcolor[HTML]{EEFFDD}{41.1} & \cellcolor[HTML]{EEFFDD}{62.4} & \multicolumn{1}{c|}{\cellcolor[HTML]{EEFFDD}{59.8}}
& \cellcolor[HTML]{EEFFDD}{71.3} & \cellcolor[HTML]{EEFFDD}{41.1} & \cellcolor[HTML]{EEFFDD}{62.4} & \multicolumn{1}{c|}{\cellcolor[HTML]{EEFFDD}{59.8}}
& \cellcolor[HTML]{EEFFDD}{71.3} & \cellcolor[HTML]{EEFFDD}{41.2} & \cellcolor[HTML]{EEFFDD}{62.3} & \cellcolor[HTML]{EEFFDD}{59.8} & \cellcolor[HTML]{EEFFDD}{58.6} \\
\multicolumn{1}{l}{\cellcolor[HTML]{EEFFDD}{~~~~~~~~~~~~~\textit{+ SDA}}} & \cellcolor[HTML]{EEFFDD}{}
& \cellcolor[HTML]{EEFFDD}{72.0} & \cellcolor[HTML]{EEFFDD}{45.4} & \cellcolor[HTML]{EEFFDD}{63.7} & \multicolumn{1}{c|}{\cellcolor[HTML]{EEFFDD}{60.0}}
& \cellcolor[HTML]{EEFFDD}{71.7} & \cellcolor[HTML]{EEFFDD}{45.4} & \cellcolor[HTML]{EEFFDD}{63.6} & \multicolumn{1}{c|}{\cellcolor[HTML]{EEFFDD}{60.1}}
& \cellcolor[HTML]{EEFFDD}{71.8} & \cellcolor[HTML]{EEFFDD}{45.4} & \cellcolor[HTML]{EEFFDD}{63.7} & \multicolumn{1}{c|}{\cellcolor[HTML]{EEFFDD}{60.1}}
& \cellcolor[HTML]{EEFFDD}{71.7} & \cellcolor[HTML]{EEFFDD}{45.3} & \cellcolor[HTML]{EEFFDD}{63.6} & \cellcolor[HTML]{EEFFDD}{60.0} & \cellcolor[HTML]{EEFFDD}{60.2} \\ 

\multicolumn{1}{l}{\cellcolor[HTML]{DDEECC}{BECoTTA (Ours)-M}} &  \cellcolor[HTML]{DDEECC}{}
& \cellcolor[HTML]{DDEECC}{72.3} & \cellcolor[HTML]{DDEECC}{42.0} & \cellcolor[HTML]{DDEECC}{63.5} & \multicolumn{1}{c|}{\cellcolor[HTML]{DDEECC}{60.1}} 
& \cellcolor[HTML]{DDEECC}{72.3} & \cellcolor[HTML]{DDEECC}{41.9} & \cellcolor[HTML]{DDEECC}{63.6} & \multicolumn{1}{c|}{\cellcolor[HTML]{DDEECC}{60.2}} 
& \cellcolor[HTML]{DDEECC}{72.3} & \cellcolor[HTML]{DDEECC}{41.9} & \cellcolor[HTML]{DDEECC}{63.6} & \multicolumn{1}{c|}{\cellcolor[HTML]{DDEECC}{60.3}} 
& \textbf{\cellcolor[HTML]{DDEECC}{72.3}} & \cellcolor[HTML]{DDEECC}{41.9} & \cellcolor[HTML]{DDEECC}{63.5} & \cellcolor[HTML]{DDEECC}{60.2} & \cellcolor[HTML]{DDEECC}{59.4} \\
\multicolumn{1}{l}{\cellcolor[HTML]{DDEECC}{~~~~~~~~~~~~~\textit{+ SDA}}} & \cellcolor[HTML]{DDEECC}{} 
& \cellcolor[HTML]{DDEECC}{71.8} & \cellcolor[HTML]{DDEECC}{48.0} & \cellcolor[HTML]{DDEECC}{66.3} & \multicolumn{1}{c|}{\cellcolor[HTML]{DDEECC}{62.0}} 
& \cellcolor[HTML]{DDEECC}{71.8} & \cellcolor[HTML]{DDEECC}{47.7} & \cellcolor[HTML]{DDEECC}{66.3} & \multicolumn{1}{c|}{\cellcolor[HTML]{DDEECC}{61.9}} 
& \cellcolor[HTML]{DDEECC}{71.8} & \cellcolor[HTML]{DDEECC}{47.8} & \cellcolor[HTML]{DDEECC}{\textbf{66.4}} & \multicolumn{1}{c|}{\cellcolor[HTML]{DDEECC}{61.9}} 
& \cellcolor[HTML]{DDEECC}{71.8} & \cellcolor[HTML]{DDEECC}{47.9} & \cellcolor[HTML]{DDEECC}{\textbf{66.3}} & \cellcolor[HTML]{DDEECC}{62.6} & \cellcolor[HTML]{DDEECC}{62.0} \\ 

\multicolumn{1}{l}{\cellcolor[HTML]{CCDDBB}{BECoTTA (Ours)-L}} &\cellcolor[HTML]{CCDDBB}{}
& \cellcolor[HTML]{CCDDBB}{71.5} & \cellcolor[HTML]{CCDDBB}{42.6} & \cellcolor[HTML]{CCDDBB}{63.2} & \multicolumn{1}{c|}{\cellcolor[HTML]{CCDDBB}{59.1}}
& \cellcolor[HTML]{CCDDBB}{71.5} & \cellcolor[HTML]{CCDDBB}{42.5} & \cellcolor[HTML]{CCDDBB}{63.2} & \multicolumn{1}{c|}{\cellcolor[HTML]{CCDDBB}{59.1}}
& \cellcolor[HTML]{CCDDBB}{71.5} & \cellcolor[HTML]{CCDDBB}{42.5} & \cellcolor[HTML]{CCDDBB}{63.2} & \multicolumn{1}{c|}{\cellcolor[HTML]{CCDDBB}{59.1}}
& \cellcolor[HTML]{CCDDBB}{71.6} & \cellcolor[HTML]{CCDDBB}{42.5} & \cellcolor[HTML]{CCDDBB}{63.1} & \cellcolor[HTML]{CCDDBB}{59.1} & \cellcolor[HTML]{CCDDBB}{59.1} \\
\multicolumn{1}{l}{\cellcolor[HTML]{CCDDBB}{~~~~~~~~~~~~~\textit{+ SDA}}} &
\cellcolor[HTML]{CCDDBB}{} & \cellcolor[HTML]{CCDDBB}{\textbf{72.7}} & \cellcolor[HTML]{CCDDBB}{\textbf{49.5}} & \cellcolor[HTML]{D3D3D3}{\textbf{66.3}} & \multicolumn{1}{c|}{\cellcolor[HTML]{CCDDBB}{\textbf{63.1}}}
& \cellcolor[HTML]{CCDDBB}{\textbf{72.5}} & \cellcolor[HTML]{CCDDBB}{\textbf{49.7}} & \cellcolor[HTML]{CCDDBB}{\textbf{66.2}} & \multicolumn{1}{c|}{\cellcolor[HTML]{CCDDBB}{\textbf{63.1}}}
& \cellcolor[HTML]{CCDDBB}{\textbf{72.3}} & \cellcolor[HTML]{CCDDBB}{\textbf{49.5}} & \cellcolor[HTML]{CCDDBB}{66.2} & \multicolumn{1}{c|}{\cellcolor[HTML]{CCDDBB}{\textbf{63.1}}}
& \cellcolor[HTML]{CCDDBB}{72.1} & \cellcolor[HTML]{CCDDBB}{\textbf{49.2}} & \cellcolor[HTML]{CCDDBB}{66.2} & \cellcolor[HTML]{CCDDBB}{\textbf{63.2}} & \cellcolor[HTML]{CCDDBB}{\textbf{63.0}} \\ \hline

\end{tabular}
}
\label{table:main1_ten}
\end{table*}

\input{table/ten_table}


\begin{table*}[t]
\centering
\renewcommand{\arraystretch}{1.3}
\setlength{\tabcolsep}{.13em}
\caption{\textbf{Quantitative results of \textit{Continual Gradual Shifts (CGS) scenarios}.} We present the results of up to ten rounds. Our CGS exhibits relatively higher performance across all models compared to the disjoint scenario, as neighboring domains are exposed within the scenario, unlike the conventional disjoint scenario. }
{\scriptsize
\begin{tabular}{lc|cccccccccccccccc|c}
\hline
\multicolumn{2}{l|}{\textbf{Round}} & \multicolumn{4}{c|}{\textbf{1}} & \multicolumn{4}{c|}{\textbf{3}} & \multicolumn{4}{c|}{\textbf{7}} & \multicolumn{4}{c|}{\textbf{10}} & \multicolumn{1}{l}{} \\ \hline

\multicolumn{1}{l|}{Method} & Parameter & Task 1 & Task 2 & Task 3 & \multicolumn{1}{c|}{Task 4} 
& Task 1 & Task 2 & Task 3 & \multicolumn{1}{c|}{Task 4} 
& Task 1 & Task 2 & Task 3 & \multicolumn{1}{c|}{Task 4} 
& Task 1 & Task 2 & Task 3 & Task 4 & Mean \\ \hline

\multicolumn{1}{l|}{Source only} & \textit{-} 
& 57.9 & 44.1 & 55.5 & \multicolumn{1}{c|}{54.7} 
& 57.9 & 44.1 & 55.5 & \multicolumn{1}{c|}{54.7} 
& 57.9 & 44.1 & 55.5 & \multicolumn{1}{c|}{54.7} 
& 57.9 & 44.1 & 55.5 & 54.7 & 54.7 \\

\multicolumn{1}{l|}{TENT~\cite{tent}} & 0.02M & 58.1 & 44.6 & 56.3 & \multicolumn{1}{c|}{55.2} 
& 58.5 & 45.1 & 56.8 & \multicolumn{1}{c|}{54.9} 
& 56.8 & 43.3 & 54.3 & \multicolumn{1}{c|}{52.0}     
& 55.1 & 41.4 & 51.5 & 49.2 & 52.0 \\

\multicolumn{1}{l|}{SAR~\cite{sar}} & 0.02M
& 57.9 & 44.2 & 55.6 & \multicolumn{1}{c|}{54.9} 
& 58.2 & 44.4 & 56.0 & \multicolumn{1}{c|}{55.2} 
& 58.3 & 44.7 & 56.4 & \multicolumn{1}{c|}{55.5} 
& 58.4 & 44.7 & 56.5 & 55.5 & 53.5 \\

\multicolumn{1}{l|}{EcoTTA~\cite{ecotta}} & 3.46M
& 62.1 & 47.6 & 59.7 & \multicolumn{1}{c|}{58.7} 
& 61.8 & 47.6 & 59.8 & \multicolumn{1}{c|}{58.6} 
& 60.9 & 47.1 & 59.1 & \multicolumn{1}{c|}{57.9} 
& 59.8 & 46.1 & 58.2 & 57.2 & 56.3 \\ 

\multicolumn{1}{l|}{\cellcolor[HTML]{EEFFDD}{BECoTTA (Ours)-S}} & \cellcolor[HTML]{EEFFDD}{0.09M}
& \cellcolor[HTML]{EEFFDD}{61.8} & \cellcolor[HTML]{EEFFDD}{46.9} & \cellcolor[HTML]{EEFFDD}{57.6} & \multicolumn{1}{c|}{\cellcolor[HTML]{EEFFDD}{56.9}} 
& \cellcolor[HTML]{EEFFDD}{61.8} & \cellcolor[HTML]{EEFFDD}{46.9} & \cellcolor[HTML]{EEFFDD}{57.6} & \multicolumn{1}{c|}{\cellcolor[HTML]{EEFFDD}{56.9}} 
& \cellcolor[HTML]{EEFFDD}{61.6} & \cellcolor[HTML]{EEFFDD}{49.8} & \cellcolor[HTML]{EEFFDD}{57.7} & \multicolumn{1}{c|}{\cellcolor[HTML]{EEFFDD}{57.0}} 
& \cellcolor[HTML]{EEFFDD}{61.7} & \cellcolor[HTML]{EEFFDD}{49.7} & \cellcolor[HTML]{EEFFDD}{57.5} & \cellcolor[HTML]{EEFFDD}{57.1} & \cellcolor[HTML]{EEFFDD}{55.8} \\
\multicolumn{1}{l|}{\cellcolor[HTML]{EEFFDD}{~~~~~~~~~~~~~\textit{+ SDA}}} & \cellcolor[HTML]{EEFFDD}{0.12M}
& \cellcolor[HTML]{EEFFDD}{62.0} & \cellcolor[HTML]{EEFFDD}{51.0} & \cellcolor[HTML]{EEFFDD}{59.9} & \multicolumn{1}{c|}{\cellcolor[HTML]{EEFFDD}{57.7}} 
& \cellcolor[HTML]{EEFFDD}{62.0} & \cellcolor[HTML]{EEFFDD}{51.0} & \cellcolor[HTML]{EEFFDD}{59.8} & \multicolumn{1}{c|}{\cellcolor[HTML]{EEFFDD}{57.6}} 
& \cellcolor[HTML]{EEFFDD}{62.0} & \cellcolor[HTML]{EEFFDD}{51.0} & \cellcolor[HTML]{EEFFDD}{59.8} & \multicolumn{1}{c|}{\cellcolor[HTML]{EEFFDD}{57.6}} 
& \cellcolor[HTML]{EEFFDD}{62.2} & \cellcolor[HTML]{EEFFDD}{51.0} & \cellcolor[HTML]{EEFFDD}{59.6} & \cellcolor[HTML]{EEFFDD}{57.8} & \cellcolor[HTML]{EEFFDD}{55.9} \\

\multicolumn{1}{l|}{\cellcolor[HTML]{DDEECC}{BECoTTA (Ours)-M}} &  \cellcolor[HTML]{DDEECC}{0.63M}
& \cellcolor[HTML]{DDEECC}{60.4} & \cellcolor[HTML]{DDEECC}{46.2} & \cellcolor[HTML]{DDEECC}{58.2} & \multicolumn{1}{c|}{\cellcolor[HTML]{DDEECC}{57.4}} 
& \cellcolor[HTML]{DDEECC}{60.5} & \cellcolor[HTML]{DDEECC}{46.2} & \cellcolor[HTML]{DDEECC}{58.2} & \multicolumn{1}{c|}{\cellcolor[HTML]{DDEECC}{57.4}} 
& \cellcolor[HTML]{DDEECC}{60.5} & \cellcolor[HTML]{DDEECC}{46.2} & \cellcolor[HTML]{DDEECC}{58.2} & \multicolumn{1}{c|}{\cellcolor[HTML]{DDEECC}{57.5}} 
& \cellcolor[HTML]{DDEECC}{60.5} & \cellcolor[HTML]{DDEECC}{46.2} & \cellcolor[HTML]{DDEECC}{58.2} & \cellcolor[HTML]{DDEECC}{57.4} & \cellcolor[HTML]{DDEECC}{55.5} \\
\multicolumn{1}{l|}{\cellcolor[HTML]{DDEECC}{~~~~~~~~~~~~~\textit{+ SDA}}} & \cellcolor[HTML]{DDEECC}{0.77M}
& \cellcolor[HTML]{DDEECC}{64.0} & \cellcolor[HTML]{DDEECC}{53.2} & \cellcolor[HTML]{DDEECC}{60.6} & \multicolumn{1}{c|}{\cellcolor[HTML]{DDEECC}{58.5}} 
& \cellcolor[HTML]{DDEECC}{63.9} & \cellcolor[HTML]{DDEECC}{53.3} & \cellcolor[HTML]{DDEECC}{60.6} & \multicolumn{1}{c|}{\cellcolor[HTML]{DDEECC}{58.6}} 
& \cellcolor[HTML]{DDEECC}{63.9} & \cellcolor[HTML]{DDEECC}{53.3} & \cellcolor[HTML]{DDEECC}{60.7} & \multicolumn{1}{c|}{\cellcolor[HTML]{DDEECC}{58.5}} 
& \cellcolor[HTML]{DDEECC}{64.0} & \cellcolor[HTML]{DDEECC}{53.3} & \cellcolor[HTML]{DDEECC}{60.7} & \cellcolor[HTML]{DDEECC}{58.5} & \cellcolor[HTML]{DDEECC}{59.0} \\ 

\multicolumn{1}{l|}{\cellcolor[HTML]{CCDDBB}{BECoTTA (Ours)-L} } &   \cellcolor[HTML]{CCDDBB}{3.16M}
& \cellcolor[HTML]{CCDDBB}{62.5} & \cellcolor[HTML]{CCDDBB}{47.7} & \cellcolor[HTML]{CCDDBB}{59.3} & \multicolumn{1}{c|}{\cellcolor[HTML]{CCDDBB}{59.0}} 
& \cellcolor[HTML]{CCDDBB}{62.5} & \cellcolor[HTML]{CCDDBB}{47.7} & \cellcolor[HTML]{CCDDBB}{59.3} & \multicolumn{1}{c|}{\cellcolor[HTML]{CCDDBB}{59.0}} 
& \cellcolor[HTML]{CCDDBB}{62.4} & \cellcolor[HTML]{CCDDBB}{47.8} & \cellcolor[HTML]{CCDDBB}{59.2} & \multicolumn{1}{c|}{\cellcolor[HTML]{CCDDBB}{58.8}} 
& \cellcolor[HTML]{CCDDBB}{62.5} & \cellcolor[HTML]{CCDDBB}{47.7} & \cellcolor[HTML]{CCDDBB}{59.1} & \cellcolor[HTML]{CCDDBB}{58.8} & \cellcolor[HTML]{CCDDBB}{57.1} \\
\multicolumn{1}{l|}{\cellcolor[HTML]{CCDDBB}{~~~~~~~~~~~~~\textit{+ SDA}}} &  \cellcolor[HTML]{CCDDBB}{\textbf{3.31M}}
& \cellcolor[HTML]{CCDDBB}{\textbf{64.6}} & \cellcolor[HTML]{CCDDBB}{\textbf{53.5}} & \cellcolor[HTML]{CCDDBB}{\textbf{62.5}} & \multicolumn{1}{c|}{\cellcolor[HTML]{CCDDBB}{\textbf{60.1}}} 
& \cellcolor[HTML]{CCDDBB}{\textbf{64.7}} & \cellcolor[HTML]{CCDDBB}{\textbf{53.6}} & \cellcolor[HTML]{CCDDBB}{\textbf{62.5}} & \multicolumn{1}{c|}{\cellcolor[HTML]{CCDDBB}{\textbf{60.1}}} 
& \cellcolor[HTML]{CCDDBB}{\textbf{64.6}} & \cellcolor[HTML]{CCDDBB}{\textbf{53.5}} & \cellcolor[HTML]{CCDDBB}{\textbf{62.4}} & \multicolumn{1}{c|}{\cellcolor[HTML]{CCDDBB}{\textbf{60.2}}} 
& \cellcolor[HTML]{CCDDBB}{\textbf{64.5}} & \cellcolor[HTML]{CCDDBB}{\textbf{53.4}} & \cellcolor[HTML]{CCDDBB}{\textbf{62.3}} & \cellcolor[HTML]{CCDDBB}{\textbf{60.1}} & \cellcolor[HTML]{CCDDBB}{\textbf{60.2}} \\ \hline

\end{tabular}
}%

\label{table:blurry_ten}
\end{table*}

\begin{table*}[!ht]
    \centering
    \renewcommand{\arraystretch}{1.3}
    \setlength{\tabcolsep}{.33em}
    \caption{\textbf{Classification error rate (\%) for CIFAR10-to-CIFAR10C with severity level 5}. Results are evaluated on WideResNet-28. * indicates our implemented version performances. }
    {\scriptsize
    \begin{tabular}{l|ccccccccccccccc|c}
    \hline
        Method & Gaus. & Shot & Impu. & Defo. & Glas. & Moti. & Zoom & Snow & Fros. & Fog & Brig. & Cont. & Elas. & Pixe. & Jpeg & Avg. err \\ \hline 
        Source & 72.3 & 65.7 & 72.9 & 46.9 & 54.3 & 34.8 & 42.0 & 25.1 & 41.3 & 26.0 & 9.3 & 46.7 & 26.6 & 58.5 & 30.3 & 43.5 \\ 
        tBN~\cite{tbn} & 28.6 & 26.8 & 37.0 & 13.2 & 35.4 & 14.4 & 12.6 & 18.0 & 18.2 & 16.0 & 8.6 & 13.3 & 24.0 & 20.3 & 27.8 & 20.9 \\
        Single do. TENT~\cite{tent} & 25.2 & 23.8 & 33.5 & 12.8 & 32.3 & 14.1 & 11.7 & 16.4 & 17.0 & 14.4 & 8.4 & 12.2 & 22.8 & 18.0 & 24.8 & 19.2 \\ 
        Continual TENT~\cite{tent} & 25.2 & 20.8 & 29.4 & 14.4 & 31.5 & 15.4 & 14.2 & 18.8 & 17.5 & 17.3 & 10.9 & 14.9 & 23.6 & 20.2 & 25.6 & 20.0 \\ 
        TTT++~\cite{ttt} & 27.9 & 25.8 & 35.8 & 13.0 & 34.3 & 14.2 & 12.2 & 17.4 & 17.6 & 15.5 & 8.6 & 13.1 & 23.1 & 19.6 & 26.6 & 20.3 \\ 
        SWRNSP~\cite{swr} & 24.6 & 20.5 & 29.3 & 12.4 & 31.1 & 13.0 & 11.3 & 15.3 & 14.7 & 11.7 & 7.8 & 9.3 & 21.5 & 15.6 & 20.3 & 17.2 \\
        NOTE~\cite{note} & 30.4 & 26.7 & 34.6 & 13.6 & 36.3 & 13.7 & 13.9 & 17.2 & 15.8 & 15.2 & 9.1 & 7.5 & 24.1 & 18.4 & 25.9 & 20.2 \\ 
        EATA~\cite{eata} & 23.8 & 18.8 & 27.3 & 13.9 & 29.7 & 16.0 & 13.3 & 18.0 & 16.9 & 15.7 & 10.5 & 12.2 & 22.9 & 17.1 & 23.0 & 18.6 \\ 
        CoTTA~\cite{cotta} & 24.3 & 21.6 & 26.6 & 11.6 & 27.6 & 12.2 & 10.3 & 14.8 & 14.1 & 12.4 & 7.5 & 10.6 & 18.3 & 13.4 & 17.3 & 16.2 \\ 
        CoTTA* & 24.6 & 21.6 & 26.5 & 12.1 & 28.0 & 13.0 & 10.9 & 15.3 & 14.6 & 13.6 & 8.1 & 12.2 & 20.0 & 14.9 & 19.5 & 17.0 \\ 
        EcoTTA (k=4)~\cite{ecotta} & 23.5 & 19.0 & 26.6 & 11.5 & 28.1 & 13.1 & 10.9 & 15.2 & 14.5 & 13.1 & 7.8 & 11.4 & 20.9 & 15.4 & 20.8 & 16.9 \\ 
        EcoTTA (k=4)* & 25.7 & 21.5 & 28.4 & 11.4 & 31.0 & 14.1 & 11.9 & 16.7 & 15.3 & 13.9 & 8.9 & 12.4 & 20.4 & 16.1 & 20.7 & 17.9 \\ \hline
        BECoTTA (w/o SDA) & \textbf{22.9} & 19.1 & 26.9 & \textbf{10.2} & \textbf{27.5} & \textbf{12.7} & \textbf{10.4} & \textbf{14.7} & \textbf{14.3} & 12.4 & \textbf{7.2} & 9.4 & 20.9 & 15.2 & 20.2 & \textbf{16.3} \\ \hline
    \end{tabular}
    }
\label{table:cifar10}
\end{table*}


\begin{table}[h]
    \centering
    \caption{\textbf{Classification error rate (\%) for standard CIFAR10-to-CIFAR10C with inference time and memory consumption (MB)}.}  
    \label{table:cifar10effi}
    \renewcommand{\arraystretch}{1.3}
    {\footnotesize
    \begin{tabular}{l|ccc}
    \hline
        Method & Avg. Err. & Time (s) & Memory (MB) \\ \hline
        TENT~\cite{tent} & 20.0 & 1222 & 118.2 \\ 
        CoTTA~\cite{cotta} & 17.0 & 18877 & 537.0 \\ 
        EcoTTA*~\cite{ecotta} & 17.9 & 4429 & 328.3 \\
        Ours (Exp10, k4) & 16.3 & 2475 & 211.6 \\ \hline
    \end{tabular}%
    }
\end{table}

\subsection{Results on Classification Tasks}

To validate the versatility of BECoTTA, we conduct additional experiments on classification task scenarios in~\Cref{table:cifar10} and \Cref{table:cifar100cc}. We also provide the computational inference time/memory efficiency in \Cref{table:cifar10effi}. 
All of our BECoTTA are initialized without SDA, which means it ensures a fair comparison with other CTTA baselines. 
For other baselines' performances, we borrow performances from Table 19 of the EcoTTA paper.

For the CIFAR10-CIFAR10C task, we adopt the WideResNet-28 backbone. 
As shown in \Cref{table:cifar10}, our BECoTTA achieves a lower error rate by \textbf{4.2\%p} while reducing inference time by \textbf{86.8\%p} compared to CoTTA, demonstrating that BECoTTA without SDA initialization consistently shows improved performance with remarkable parameter efficiency compared to strong CTTA baselines. 
Here, we additionally measure the performance of EcoTTA based on the community re-implementation version\footnote{https://github.com/Lily-Le/EcoTTA} since the official code and checkpoints are not public.

For the CIFAR100-CIFAR100C task, we adopt the WideResNet-40 backbone. 
While CIFAR100 is a larger scale dataset than CIFAR10, as shown in \Cref{table:cifar100cc}, BECoTTA outperforms all of the other CTTA baselines in most of the sections.

\subsection{Additional Ablations}

\begin{table}[h]
\centering
\caption{\textbf{Ablation study about warm-up loss weights}. While doing this ablation, we set the hidden dimension as [8,8,16,32] and utilize six experts with $k$=3.}   
\label{table:loss_weight}
\renewcommand{\arraystretch}{1.2}
{\small
\begin{tabular}{cc|ccccc|c}
\hline
$\bm{\lambda_s}$ & $\bm{\lambda_m}$ & B-Clear & A-Fog & A-Night & A-Snow & B-Over & Avg \\ \hline
0.5 & 0.5 & 45.35 & \textbf{70.92} & 43.17 & \textbf{59.59} & 50.56 & \textbf{53.92} \\
0.5 & 0.01 & 45.2 & 70.61 & \textbf{43.58} & 59.31 & 50.48 & 53.84 \\
1 & 0.5 & \textbf{45.47} & 69.84 & 43.29 & 59.28 & \textbf{50.82} & 53.74 \\
1 & 0.001 & 45.38 & 70.26 & 42.66 & 58.79 & 50.51 & 53.52 \\
5 & 0.01 & 45.02 & 70.24 & 42.57 & 59.07 & 50.42 & 53.46 \\ \hline
\end{tabular}%
}
\end{table}

\textbf{More about the loss weights.}
To assess the impact of warm-up loss weight $\lambda_s$ and $\lambda_m$, we conduct an ablation study in which $\lambda_s$ is the segmentation loss weight and $\lambda_m$ is the mutual loss weight in ~\Cref{table:loss_weight}. We fix the domain discriminator loss weight $\lambda_d$ while doing ablations. 
To precisely measure the effects, we evaluate the zero-shot performance for each domain, excluding the TTA process. 
The results indicate that as the weight of the mutual loss decreases, the performance of night scenes increases as it relatively diminishes consideration for mutual information. 
Furthermore, similar trends in performance are observed for adjusting the loss weight of similar images, such as \{\textit{BDD-Clear, BDD-Overcast}\} and \{\textit{ACDC-Fog, ACDC-Snow}\}.

\textbf{More about the routing policy.} 
We also conduct ablation studies on the routing policy for selecting experts within the MoDE layer. 
We measure the impact of the routing policy in the \textit{CDS-Hard} scenario under the \textit{w/ SDA} setting, where [2, 4, 10, 16] hidden dimensions and six experts with $k$=3.  
The multi-task performance refers to using a fixed assignment per domain, and stochastic routing~\cite{thor, adamix} involves random-wise selection. 
According to ~\Cref{tab:routing}, our chosen top-k routing demonstrates the best performance.
This is because the domain-specific router allows for routing that takes input-wise information into consideration.

\begin{table}[h]
        \centering
    \caption{\textbf{Ablation study for each routing policy.} We conduct routing policy ablation using hidden dimension [2,4,10,16] with six experts.}
        \renewcommand{\arraystretch}{1.3}
        {\footnotesize
            \begin{tabular}{l|ccccc|c}
            \hline 
            \multicolumn{1}{c|}{} & B-Clear & A-Fog & A-Night & A-Snow & B-Overcast & Avg \\ \hline
            Multi-task & 44.56 & 68.99 & 37.66 & 58.59 & 50.14 & 52.00\\
            Stochastic & 45.40 & 69.74 & \textbf{42.85} & 58.81 & 50.65 & 53.50 \\
            \textbf{Top-K(Ours)} & \textbf{45.54}  & \textbf{70.77} & 42.62  & \textbf{59.66} & \textbf{50.76} & \textbf{53.87} \\ \hline
            \end{tabular}%
        }%
        \vspace{.3cm}
        \label{tab:routing}
     \label{tab:ablations}
\end{table}

\textbf{More about hidden dimension.}
We include results considering higher hidden dimensions and the number of experts in ~\Cref{table:hiddendim}, along with the consumption of parameters and memory. 
The hidden dimension refers to the rank $r$ for each encoder stage block for Segformer~\cite{segformer}. 
For instance, [2,4,10,16] means each $r$=2,4,10,16 used at the MoDE layer for four stages Segformer~\cite{segformer}. 
[0,0,0,16] denotes that the MoDE layer is used only in the last stage of the encoder. 
In particular, we predominantly opt for relatively low hidden dimensions and fewer experts, considering the trade-off with efficiency, even though setting a higher hidden dimension generally ensures better performance with more parameters.


\subsection{Metrics for Continual Learning} 
Both CTTA and continual learning share the common objective of preventing forgetting to retain information encountered in the online stream. 
Therefore, we adopt the continual learning metrics (AvgIoU, BWT) for evaluating the forgetting phenomenon as represented in ~\Cref{table:cl_metrics_v2}. 

AvgIoU denotes the overall performance while doing the learning process, and BWT evaluates the average influence of the current $N$th round on all of the previous tasks. 
These two metrics at the $k$th round are commonly defined as below. 
We measure the AvgIoU and BWT in the CTTA process after each round is finished.

\begin{equation}
AvgIoU_k = \frac{1}{\mathcal{D}}\sum_{j=1}^\mathcal{D} a_{kj}
\end{equation}

\begin{equation}
BWT_k = \frac{1}{\mathcal{D}}\sum_{j=1}^\mathcal{D} a_{kj}-\tilde{a_j}
\end{equation}

where $\mathcal{D}$ is the number of domains in each round, $a_{kj}$ denotes IoU evaluated by the model trained $k$ round for the $j$th domain, and $\tilde{a_j}$ represents IoU evaluated in the $j$th domain by the model trained up to the $j$th domain within the $k$ rounds.


\input{table/forgetting_metrics}

In the case of TENT~\cite{tent}, both AvgIoU and BWT show a gradual decline as the round continues because of the severe effects of forgetting. 
However, our method addresses this forgetting effectively and shows the highest AvgIoU and BWT, especially when the effectiveness of domain-wise learning is maximized in \textit{w/SDA} settings. 
In particular, the BWT improves as the round progresses, so it is interpreted that current learning has a positive effect on the past domains as learning continues. 

\subsection{Initialization of BECoTTA}
We demonstrate that BECoTTA outperforms baselines under all three initialization policies, that is, even though w/o source domain data warm-up. In~\Cref{tab:init}, we compare \textit{(i) random} and \textit{(ii) source domain} initialization with other baselines on the CDS-Hard scenario.

For \textit{(i) Random initialization}, we compare non-warm-up BECoTTA with TENT and CoTTA. We randomly initialize all weights of MoDE. 
BECoTTA, without any initialization, surpasses the performance of both CoTTA and TENT.

For \textit{(ii) Source domain initialization}, we compare source-initialized (w/o SDA) BECoTTA with EcoTTA.
We note that, in the current CTTA field, such quick warmup is entirely permissible, and many works [1-6] directly compare their methods with CoTTA \& TENT in a fair manner. We clarify it again to fully address the concern of the reviewer.
It is evident that BECoTTA consistently outperforms the best-performing CTTA baseline, EcoTTA, in both terms of IoU and efficiency (Table 1 of our submission), improving average accuracy by \textbf{+2.5\%p} (50.7\% vs. 48.2\%) while using fewer trainable parameters (0.09M vs. 3.46M).

\begin{table}[!ht]
    \centering
    \caption{\textbf{Ablation study for the initialization policy.}}
    \renewcommand{\arraystretch}{1.3}
    {\footnotesize
    \begin{tabular}{l|c|ccccc|c|l}
    \hline
        Model & Source Warmup & BC & AF & AN & AS & BO & Avg & Params \\ \hline
        TENT & ~ & 30.9 & 51.5 & 20.4 & 37.0 & 33.0 & 34.5 & 0.02M \\ 
        CoTTA & ~ & 43.3 & 67.3 & 34.8 & 56.9 & 48.8 & 50.2 & 54.72M \\ 
        Ours-S & ~ & \textbf{43.4} & \textbf{67.7} & \textbf{35.0} & \textbf{57.3} & \textbf{49.0} & \textbf{50.4} & 0.09M \\ 
        Ours-M & ~ & \textbf{43.4} & 67.6 & 35.0 & 57.2 & 49.1 & 50.4 & 0.63M \\ 
        Ours-L & ~ & 43.1 & 67.6 & 35.0 & 57.0 & 48.5 & 50.2 & 3.16M \\ \hline
        EcoTTA & \ding{52} & 41.9 & 66.1 & 31.5 & 55.3 & 46.2 & 48.2 & 3.46M \\ 
        Ours-S & \ding{52} & 43.0 & \textbf{69.5} & 35.1 & 57.3 & 48.8 & 50.7 & 0.09M \\ 
        Ours-M & \ding{52} & 43.7 & 68.8 & 34.5 & 57.9 & 49.2 & 50.8 & 0.63M \\ 
        Ours-L & \ding{52} & \textbf{44.0} & 69.1 & \textbf{35.1} & \textbf{58.3} & \textbf{50.2} & \textbf{51.3} & 3.16M \\ \hline
    \end{tabular}
    }
    \label{tab:init}
\end{table}

\subsection{Comparison of Standard Deviation}
We provide the average performance over five independent runs to investigate performance fluctuation. In~\Cref{tab:sd}, BECoTTA shows the smallest standard deviation, ensuring stable performance over other baselines.

\begin{table}[!ht]
    \centering
    \caption{\textbf{Result on standard deviation on CDS-Hard scenario.} }
    \renewcommand{\arraystretch}{1.5}
    \footnotesize{
    \begin{tabular}{l|ccccc|c|c}
    \hline
        ~ & B-Clear & A-Fog & A-Night & A-Snow & B-Overcast & Avg & Parameter\\ \hline
        TENT & 41.0 ± 0.02 & 64.6 ± 0.19 & 33.3 ± 0.13 & 54.4 ± 0.12 & 46.5 ± 0.20 & 48.0 ± 0.04 & 0.02M \\ 
        CoTTA & 43.2 ± 0.06 & 67.2 ± 0.01 & 34.8 ± 0.06 & 56.9 ± 0.07 & 48.6 ± 0.11 & 50.1 ± 0.06 & 54.72M \\ 
        Ours-S (w/o SDA) & \textbf{42.9 ± 0.04} & \textbf{69.5 ± 0.01} & \textbf{35.0 ± 0.04} & \textbf{57.24 ± 0.03} & \textbf{47.8 ± 0.00} & \textbf{50.5 ± 0.02} & \textbf{0.09M} \\  \hline
    \end{tabular}
    }
    \label{tab:sd}
\end{table}

\subsection{Comparison of Inference Speed}

We additionally measure the inference time while deploying each CDS-Hard target domain during CTTA. For transparency and reliability, we evaluate the inference time per each small section of CDS-Hard. (The time difference across each domain is due to the varying number of data within each domain.) 
Our BECoTTA implements the without SDA version for a fair comparison with other baselines. 
As shown in~\Cref{tab:time}, Ours-S achieves \textbf{80.4\%p} decreased inference time, but \textbf{1.0\%p} increased performance than CoTTA.
In addition, we conducted classification experiments on the CIFAR10 - CIFAR10C dataset based on the WideResNet-28 backbone. As shown in~\Cref{table:cifar10effi}, BECoTTA achieves a lower error rate by \textbf{4.2\%p} while reducing inference time by 
\textbf{86.8\%p} compared to CoTTA.

\begin{table}[!ht]
    \centering
    \caption{\textbf{Result of the inference time on CDS-Hard scenario.} }
    \renewcommand{\arraystretch}{1.3}
    \begin{tabular}{l|ccccc|c}
    \hline
        ~ & BC & AF & AN & AS & BO & Time Avg \\ \hline
        TENT & 302.7 & 378.3 & 385.3 & 377.5 & 197.3 & 328.2 \\ 
        CoTTA & 2746.2 & 6251.6 & 6311.9 & 6239.4 & 1325.8 & 4574.9 \\ 
        EcoTTA & 2159.2 & 2159.2 & 2187.8 & 2163.6 & 708.7 & 1875.7 \\ \hline
        Ours-S & 638.1 & 1142.3 & 1154.4 & 1139.5 & 415.8 & 898.0 \\ 
        Ours-M & 967.0 & 1486.4 & 1449.0 & 1441.0 & 636.7 & 1196.0 \\ 
        Ours-L & 1229.6 & 1584.6 & 1667.7 & 1749.5 & 893.6 & 1425.0 \\ \hline 
    \end{tabular}
    \label{tab:time}
\end{table}

\subsection{Quality of SDA.} 
We adopt different augmentation methods to build realistic SDA. 
We utilize TSIT~\cite{jiang2020tsit} for style-transfer and PyTorch transformation (e.g., ColorJitter, RandomGrayscale), same as EcoTTA~\cite{ecotta}. 
As shown in \Cref{table:wad_ablation}, we verify that the quality of SDA is a less important factor to have an effect on our domain-adaptive architecture. 
This demonstrates that our BECoTTA is implemented with various augmentations, showing its potential for expansion in diverse situations.

\begin{table}[t]
\begin{minipage}[c]{\linewidth}
\centering 
\caption{\textbf{Ablation of the quality of SDA.} Our SDA has versatility with diverse augmentation methods.}
\vspace{-0.05in}
\renewcommand{\arraystretch}{1.3} 
    \footnotesize{
    \begin{tabular}{cl|cccccc}
    \hline
    \multicolumn{2}{c|}{\multirow{2}{*}{SDA Augmentation}} & \multicolumn{6}{c}{Round 1} \\  \cline{3-8} 
    \multicolumn{2}{c|}{} & B-Clear & A-Fog & A-Night & A-Snow & \multicolumn{1}{c|}{B-Over} & Avg \\ \hline
    \multicolumn{2}{c|}{BECoTTA (Ours) - M} & 43.8 & 68.8 & 34.9 & 57.9 & \multicolumn{1}{c|}{49.2} & 50.9 \\
    \multicolumn{2}{c|}{~~~~~~+ \textit{Style-transfer}} & \textbf{45.6} & \textbf{70.8} & \textbf{42.6} & \textbf{59.7} & \multicolumn{1}{c|}{\textbf{50.8}} & \textbf{53.9} \\
    \multicolumn{2}{c|}{~~~~~~+ \textit{Transformation}} & 45.3 & 70 & 43.2 & 59.5 & \multicolumn{1}{c|}{50.7} & 53.7 \\ \hline
    \end{tabular}%
    }
    \label{table:wad_ablation}
    \vspace{-0.1in}
\end{minipage}
\end{table}
\section{Dataset Construction}\label{append:data}
\subsection{Scenario Construction Process}


\textbf{CDS-\textit{Easy} scenario.}
As we mention in the main paper, we adopt the weather shift scenario in CTTA from CoTTA~\cite{cotta} for a fair comparison. 
We set the target domain using the training set of the ACDC dataset, so the dataset for each domain consists of 400 unlabeled images, and their training order is as follows: $\{\textit{Fog} \rightarrow \textit{Night} \rightarrow  \textit{Rain} \rightarrow  \textit{Snow}\}$.

\textbf{CDS-\textit{Hard} scenario.}
To incorporate domain shift based on geographical factors and weather shifts from the Cityscapes-ACDC setting, we add \textit{clear} and \textit{overcast} datasets from BDD-100k as mentioned in the paper. 
We parse the official annotation \texttt{{json}} file\footnote{\texttt{bdd100klabels\_images\_train.json}} to split the BDD-100k train dataset by weather conditions. 
For future reproducibility, we will publicly share the file list of our scenario.
Consequently, we obtain a scenario sequence of $\{\textit{BDD-Clear} \rightarrow \textit{ACDC-Fog} \rightarrow \textit{ACDC-Night} \rightarrow \textit{ACDC-Snow} \rightarrow \textit{BDD-Overcast} \}$.
Each of them consists of 500 unlabeled images. (We additionally add the ACDC 100 validation dataset together.) 

\textbf{Continual Gradual Shifts (CGS) scenario.}
To construct gradually changing weather scenarios with blurry boundaries, we first conduct sampling from a Gaussian distribution with CDS-\textit{Easy} target domains. 
Given a total of 1600 (400x4) timesteps in one round (including four tasks) at CDS-\textit{Easy}, we define sampling distributions \textbf{$\bm{\theta_{i}} \sim \mathcal{N}(400i, 200)$} for each domain $i$, and perform uniform sampling to represent gradual changes of weathers. In the end, we construct four tasks containing blurry boundaries of weather as illustrated in \Cref{fig:overall}.

\subsection{Source Dataset Augmentation (SDA)}

\begin{figure}[h]
    \centering
    {\includegraphics[width=.7\linewidth] {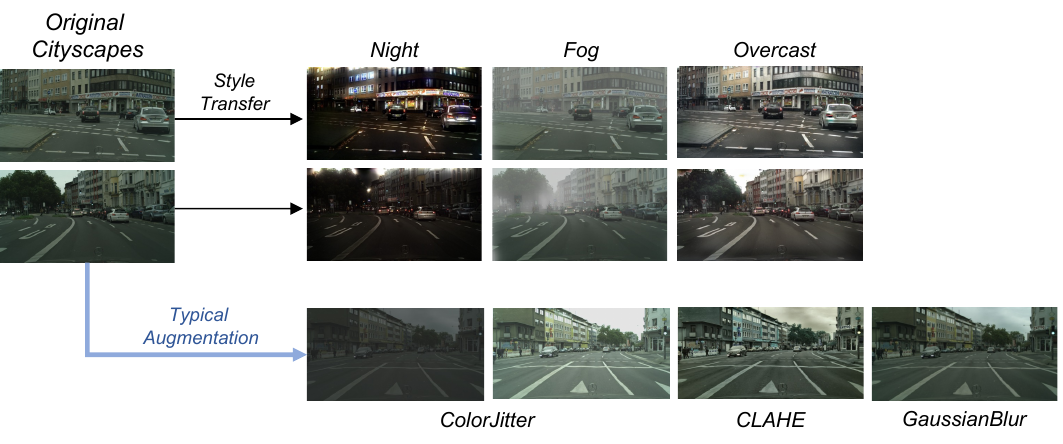}} 
    \caption{\textbf{The example of generating SDA with different augmentation.} We apply pre-trained style transfer and PyTorch augmentation for generating candidate domains for SDA.}
    \label{fig:warmup_process}
\end{figure}

\textbf{Generating process.} 
We utilize the pre-trained style transformer TSIT~\cite{jiang2020tsit} to generate candidate domains using the Cityscapes~\cite{cityscapes}. 
For the candidate domains, we set dark, bright, and foggy styles to represent real-world weather practically as illustrated in ~\Cref{fig:warmup_process}. 
We also apply the simple PyTorch augmentation to recreate them. 
Note that this process does not involve any training steps and resembles a one-time operation when setting the source domain. 
During the warm-up process, it enables the initialization from pre-defined domains by updating only the domain-wise routers and experts of the MoDE layer.
Moreover, we have the flexibility to freely expand these candidate domains to others.

\section{Limitation}
We verify our BECoTTA demonstrates superior performance with fewer parameters compared to other CTTA baselines. However, it requires the user to choose from a range of hyperparameters, including the hidden dimension of experts (\textit{dim}), the number of experts ($N$), and the number of selected experts ($K$). This variability sometimes leads to slight performance fluctuations, but it also highlights the flexibility of BECoTTA. Through extensive empirical experiments, we have confirmed that BECoTTA consistently outperforms all current CTTA baselines, regardless of the hyperparameter settings.
Furthermore, while BECoTTA operates effectively with random initialization, its performance is significantly enhanced when optimized in conjunction with SAD, maximizing the benefits of domain knowledge transfer. 
This approach does not contradict the assumptions of CTTA, as numerous studies have permitted initialization from the source domain. However, the necessity for a warmup phase could be considered a drawback.

\input{table/full_hiddendim_ablation}




%% file: table/ten_table.tex
\begin{table*}[h]
\centering
\renewcommand{\arraystretch}{1.25}
\caption{\textbf{Results on \textit{CDS-Hard (imbalanced weather \& area shifts)}.} We devise a novel scenario encompassing imbalanced weather and area shifts. We present performance results for both \textit{w/o WAD} and \textit{w/ WAD} across the overall baselines. We report $S$, $M$, and $L$ versions for our BECoTTA based on the number of parameters. }
\vspace{-0.05in}
\setlength{\tabcolsep}{.2em}

\resizebox{\linewidth}{!}{%
\begin{tabular}{clc|cccccc|cccccc|c|c}
\toprule
\multicolumn{3}{c|}{Round} & \multicolumn{6}{c|}{\textbf{1}} & \multicolumn{6}{c|}{\textbf{4}} & \multirow{2}{*}{$\Delta$} &  Parameter \\ \cline{1-15}

\multicolumn{1}{c|}{Init} & \multicolumn{1}{c|}{Method}  & Init update & B-Clear & A-Fog & A-Night & A-Snow & \multicolumn{1}{c|}{B-Overcast} & Mean & \multicolumn{1}{l}{B-clear} & \multicolumn{1}{l}{A-Fog} & \multicolumn{1}{l}{A-Night} & \multicolumn{1}{l}{A-Snow} & \multicolumn{1}{l|}{B-Overcast} & \multicolumn{1}{c|}{Mean} & \\ \hline \hline

\multicolumn{1}{l|}{} & \multicolumn{1}{l|}{Source only}  & - &41.0 & 64.4 & 33.4 & 54.3 & \multicolumn{1}{c|}{46.3} & 47.9 & 
41.0 & 64.4 & 33.4 & 54.3 & \multicolumn{1}{c|}{46.3} & 47.9 & +0.0 & - \\ 

\multicolumn{1}{l|}{} & \multicolumn{1}{l|}{CoTTA~\cite{cotta}}  & - & 43.3 & 67.3 & 34.8 & 56.9 & \multicolumn{1}{c|}{48.8} & 50.2 
& 43.3 & 67.3 & 34.8 & 56.9 & \multicolumn{1}{c|}{48.8} & 50.2 & +0.0  & \textbf{54.72M}\\

\multicolumn{1}{l|}{} & \multicolumn{1}{l|}{TENT~\cite{tent}} & - & 41.1 & 64.9 & 33.2 & 54.3 & \multicolumn{1}{c|}{46.3} & 47.9 
& 38.6 & 62.0 & 28.3 & 49.1 & \multicolumn{1}{c|}{41.6} & 43.9 & \textcolor{red}{-4.0} & 0.02M \\
    
\multicolumn{1}{l|}{\textit{w/o SDA}} & \multicolumn{1}{l|}{SAR~\cite{sar}} &  - & 41.0 & 64.5 & 33.4 & 54.5 & \multicolumn{1}{c|}{46.6} & 48.0 & 
41.4 & 64.8 & 33.1 & 54.8 & \multicolumn{1}{c|}{46.9} & 48.2 & +0.2  & 0.02M \\
    
\multicolumn{1}{l|}{} & \multicolumn{1}{l|}{EcoTTA~\cite{ecotta}} & \textit{\textcolor{darkcerulean}{MetaNet}} & 44.1 & 69.6 & 35.3 & 58.2 & \multicolumn{1}{c|}{49.6} & 51.3 & 
44.0 & 69.2 & 34.7 & 57.9 & \multicolumn{1}{c|}{49.1} & 50.9 & \textcolor{red}{-0.4} & 3.46M\\ 

\multicolumn{1}{l|}{}& \multicolumn{1}{l|}{\cellcolor{g}BECoTTA (S)} & 
\cellcolor{g}\textit{\textcolor{darkcerulean}{MoDE}} & 
\cellcolor{g}42.9 &
\cellcolor{g}68.5 &
\cellcolor{g}35.0 &
\cellcolor{g}57.2 &
\multicolumn{1}{c|}{\cellcolor{g}47.8} &
\cellcolor{g}50.5 &
\cellcolor{g}43.0 &
\cellcolor{g}69.5 &
\cellcolor{g}35.1 &
\cellcolor{g}57.2 &
\multicolumn{1}{c|}{\cellcolor{g}47.8} &
\cellcolor{g}50.7 &
\cellcolor{g}+0.1 &
\cellcolor{g}0.09M \\

\multicolumn{1}{l|}{}& \multicolumn{1}{l|}{\cellcolor{gg}BECoTTA (M)} & 
\cellcolor{gg}\textit{\textcolor{darkcerulean}{MoDE}} & 
\cellcolor{gg}43.8 &
\cellcolor{gg}68.8 &
\cellcolor{gg}34.9 &
\cellcolor{gg}57.9 &
\multicolumn{1}{c|}{\cellcolor{gg}49.2} & 
\cellcolor{gg}50.9 & 
\cellcolor{gg}43.7 & 
\cellcolor{gg}68.9 & 
\cellcolor{gg}34.8 & 
\cellcolor{gg}57.9 & 
\multicolumn{1}{c|}{\cellcolor{gg}49.3} &
\cellcolor{gg}50.9 &
\cellcolor{gg}+0.0 & 
\cellcolor{gg}0.63M \\

\multicolumn{1}{l|}{}& \multicolumn{1}{l|}{\cellcolor{ggg}BECoTTA (L)} &
\cellcolor{ggg}\textit{\textcolor{darkcerulean}{MoDE}} & 
\cellcolor{ggg}43.9 &
\cellcolor{ggg}69.1 &
\cellcolor{ggg}35.0 &
\cellcolor{ggg}58.3 &
\multicolumn{1}{c|}{\cellcolor{ggg}50.2} &
\cellcolor{ggg}51.3 &
\cellcolor{ggg}44.0 &
\cellcolor{ggg}69.1 &
\cellcolor{ggg}35.1 &
\cellcolor{ggg}58.3 &
\multicolumn{1}{c|}{\cellcolor{ggg}50.2} &
\cellcolor{ggg}51.3 &
\cellcolor{ggg}+0.0 &
\cellcolor{ggg}3.16M \\ \hline

\multicolumn{1}{l|}{}  & \multicolumn{1}{l|}{Source only}  & \textit{\textcolor{crimson}{Full}} & 43.6 & 68.7 & 44.5 & 59.0 & \multicolumn{1}{c|}{48.7} & 52.9 & 43.6 & 68.7 & 44.5 & 59.0 & \multicolumn{1}{c|}{48.7} & 52.9 & +0.0 & -\\  

\multicolumn{1}{l|}{} & \multicolumn{1}{l|}{CoTTA~\cite{cotta}} & \textit{\textcolor{crimson}{Full}} & 46.4 & 70.6 & 45.7 & 61.2 & \multicolumn{1}{c|}{51.3} & 55.0 
& 46.1 & 70.5 & 45.6 & 61.1 & \multicolumn{1}{c|}{51.2} & 54.9 & \textcolor{red}{-0.1}  & \textbf{54.72M}\\ 

\multicolumn{1}{l|}{} & \multicolumn{1}{l|}{TENT~\cite{tent}} & \textit{\textcolor{crimson}{Full}} & 43.7 & 68.5 & 44.6 & 59.0 & \multicolumn{1}{c|}{48.3} & 52.8 
& 41.4 & 64.6 & 40.7 & 53.5 & \multicolumn{1}{c|}{44.8} & 49.0 & \textcolor{red}{-3.8} & 0.02M \\ 

\multicolumn{1}{l|}{\textit{w/ SDA}} & \multicolumn{1}{l|}{SAR~\cite{sar}}  & \textit{\textcolor{crimson}{Full}} & 43.6 & 68.6 & 44.5 & 59.1 & \multicolumn{1}{c|}{48.7} 
& 52.9 & 43.7 & 69.1 & 36.4 & 56.8 & \multicolumn{1}{c|}{48.3} & 50.8 & \textcolor{red}{-2.1} & 0.02M \\ 

\multicolumn{1}{l|}{} & \multicolumn{1}{l|}{EcoTTA~\cite{ecotta}} & \textit{\textcolor{darkcerulean}{MetaNet}} & 44.6 & 70.2 & 41.6 & 58.0 & \multicolumn{1}{c|}{49.9} & 52.9 
& 43.7 & 69.1 & 36.4 & 56.8 & \multicolumn{1}{c|}{48.3} & 50.8 & \textcolor{red}{-2.1} & 3.46M\\ 


\multicolumn{1}{l|}{} & \multicolumn{1}{l|}{\cellcolor{g}BECoTTA+ (S)} &
\cellcolor{g}\textit{\textcolor{darkcerulean}{MoDE}} & 
\cellcolor{g}44.1 &
\cellcolor{g}69.5 &
\cellcolor{g}40.1 &
\cellcolor{g}56.8 &
\multicolumn{1}{c|}{\cellcolor{g}49.1} &
\cellcolor{g}51.9 &
\cellcolor{g}44.0 &
\cellcolor{g}69.4 &
\cellcolor{g}40.2 &
\cellcolor{g}56.9 &
\multicolumn{1}{c|}{\cellcolor{g}49.2} &
\cellcolor{g}51.8 &
\cellcolor{g}+0.0 & 
\cellcolor{g}0.12M \\ 

\multicolumn{1}{l|}{} & \multicolumn{1}{l|}{\cellcolor{gg}BECoTTA+ (M) } &
\cellcolor{gg}\textit{\textcolor{darkcerulean}{MoDE}} & 
\cellcolor{gg}45.6 & 
\cellcolor{gg}70.8 & 
\cellcolor{gg}42.6 & 
\cellcolor{gg}59.6 & 
\multicolumn{1}{c|}{\cellcolor{gg}\textbf{50.8}} & 
\cellcolor{gg}53.9 & 
\cellcolor{gg}45.6 & 
\cellcolor{gg}70.8 & 
\cellcolor{gg}42.6 & 
\cellcolor{gg}59.5 & 
\multicolumn{1}{c|}{\cellcolor{gg}\textbf{50.8}} & 
\cellcolor{gg}53.9 & 
\cellcolor{gg}+0.0 & 
\cellcolor{gg}0.77M \\ 

\multicolumn{1}{l|}{} & \multicolumn{1}{l|}{\cellcolor{ggg}BECoTTA+ (L) } &
\cellcolor{ggg}\textit{\textcolor{darkcerulean}{MoDE}} & 
\cellcolor{ggg}\textbf{45.7} &
\cellcolor{ggg}\textbf{71.4} &
\cellcolor{ggg}\textbf{43.7} &
\cellcolor{ggg}\textbf{59.6} &
\multicolumn{1}{c|}{\cellcolor{ggg}50.5} &
\cellcolor{ggg}\textbf{54.2} &
\cellcolor{ggg}\textbf{45.7} &
\cellcolor{ggg}\textbf{71.3} &
\cellcolor{ggg}\textbf{43.7} &
\cellcolor{ggg}\textbf{59.6} &
\multicolumn{1}{c|}{\cellcolor{ggg}50.6} &
\cellcolor{ggg}\textbf{54.2} &
\cellcolor{ggg}\textbf{+0.0} & 
\cellcolor{ggg}3.32M \\
\bottomrule
\end{tabular}%
}

\vspace{.3cm}

\resizebox{\linewidth}{!}{%
\begin{tabular}{clc|cccccc|cccccc|c|c}
\toprule
\multicolumn{3}{c|}{Round} & \multicolumn{6}{c|}{\textbf{7}} & \multicolumn{6}{c|}{\textbf{10}} & \multirow{2}{*}{$\Delta$} &  Parameter \\ \cline{1-15}

\multicolumn{1}{c|}{Init} & \multicolumn{1}{c|}{Method}  & Init update & B-Clear & A-Fog & A-Night & A-Snow & \multicolumn{1}{c|}{B-Overcast} & Mean & \multicolumn{1}{l}{B-clear} & \multicolumn{1}{l}{A-Fog} & \multicolumn{1}{l}{A-Night} & \multicolumn{1}{l}{A-Snow} & \multicolumn{1}{l|}{B-Overcast} & \multicolumn{1}{c|}{Mean} & \\ \hline \hline

\multicolumn{1}{l|}{} & \multicolumn{1}{l|}{Source only}  & - &41.0 & 64.4 & 33.4 & 54.3 & \multicolumn{1}{c|}{46.3} & 47.9 & 41.0 & 64.4 & 33.4 & 54.3 & \multicolumn{1}{c|}{46.3} & 47.9 & +0.0 & - \\ 

\multicolumn{1}{l|}{} & \multicolumn{1}{l|}{CoTTA~\cite{cotta}}  & - & 43.3 & 67.3 & 34.8 & 56.9 & \multicolumn{1}{c|}{48.8} & 50.2 
& 43.3 & 67.3 & 34.8 & 56.9 & \multicolumn{1}{c|}{48.8} & 50.2 & +0.0  & \textbf{54.72M}\\

\multicolumn{1}{l|}{} & \multicolumn{1}{l|}{TENT~\cite{tent}} & - & 34.4 & 56.4 & 23.5 & 42.2 & \multicolumn{1}{c|}{36.7} & 38.6 
& 30.9 & 51.5 & 20.4 & 37.0 & \multicolumn{1}{c|}{33.0} & 34.6 & \textcolor{red}{-13.3} & 0.02M \\
    
\multicolumn{1}{l|}{\textit{w/o SDA}} & \multicolumn{1}{l|}{SAR~\cite{sar}} &  - & 41.4 & 64.7 & 32.4 & 54.6 & \multicolumn{1}{c|}{46.8} & 47.9 
& 41.3 & 64.3 & 31.6 & 54.2 & \multicolumn{1}{c|}{46.6} & 47.6 & \textcolor{red}{-0.4}  & 0.02M \\
    
\multicolumn{1}{l|}{} & \multicolumn{1}{l|}{EcoTTA~\cite{ecotta}} & \textit{\textcolor{darkcerulean}{MetaNet}} & 43.2 & 67.9 & 33.4 & 56.8 & \multicolumn{1}{c|}{47.9} & 49.8 
& 41.9 & 66.1 & 31.5 & 55.3 & \multicolumn{1}{c|}{46.2} & 48.2 & \textcolor{red}{-3.1} & 3.46M\\ 

\multicolumn{1}{l|}{}& \multicolumn{1}{l|}{\cellcolor{g}BECoTTA (S)} & 
\cellcolor{g}\textit{\textcolor{darkcerulean}{MoDE}} & 
\cellcolor{g}42.9 &
\cellcolor{g}69.5 &
\cellcolor{g}35.1 &
\cellcolor{g}57.3 &
\multicolumn{1}{c|}{\cellcolor{g}47.8} &
\cellcolor{g}50.5 &
\cellcolor{g}43.0 &
\cellcolor{g}69.5 &
\cellcolor{g}35.1 &
\cellcolor{g}57.3 &
\multicolumn{1}{c|}{\cellcolor{g}48.8} &
\cellcolor{g}50.7 &
\cellcolor{g}+0.1 &
\cellcolor{g}0.09M \\

\multicolumn{1}{l|}{}& \multicolumn{1}{l|}{\cellcolor{gg}BECoTTA (M)} & 
\cellcolor{gg}\textit{\textcolor{darkcerulean}{MoDE}} & 
\cellcolor{gg}43.7 &
\cellcolor{gg}68.8 &
\cellcolor{gg}34.8 &
\cellcolor{gg}57.9 &
\multicolumn{1}{c|}{\cellcolor{gg}49.1} & 
\cellcolor{gg}50.8 & 
\cellcolor{gg}43.7 & 
\cellcolor{gg}68.8 & 
\cellcolor{gg}34.5 & 
\cellcolor{gg}57.9 & 
\multicolumn{1}{c|}{\cellcolor{gg}49.2} &
\cellcolor{gg}50.9 &
\cellcolor{gg}+0.0 & 
\cellcolor{gg}0.63M \\

\multicolumn{1}{l|}{}& \multicolumn{1}{l|}{\cellcolor{ggg}BECoTTA (L)} &
\cellcolor{ggg}\textit{\textcolor{darkcerulean}{MoDE}} & 
\cellcolor{ggg}43.9 &
\cellcolor{ggg}69.0 &
\cellcolor{ggg}35.0 &
\cellcolor{ggg}58.2 &
\multicolumn{1}{c|}{\cellcolor{ggg}50.1} &
\cellcolor{ggg}51.3 &
\cellcolor{ggg}44.0 &
\cellcolor{ggg}69.1 &
\cellcolor{ggg}35.0 &
\cellcolor{ggg}58.2 &
\multicolumn{1}{c|}{\cellcolor{ggg}50.2} &
\cellcolor{ggg}51.3 &
\cellcolor{ggg}+0.0 &
\cellcolor{ggg}3.16M \\ \hline

\multicolumn{1}{l|}{}  & \multicolumn{1}{l|}{Source only}  & \textit{\textcolor{crimson}{Full}} & 43.6 & 68.7 & 44.5 & 59.0 & \multicolumn{1}{c|}{48.7} & 52.9 & 43.6 & 68.7 & 44.5 & 59.0 & \multicolumn{1}{c|}{48.7} & 52.9 & +0.0 & -\\  

\multicolumn{1}{l|}{} & \multicolumn{1}{l|}{CoTTA~\cite{cotta}} & \textit{\textcolor{crimson}{Full}} 
& 46.1 & 70.5 & 45.6 & 61.1 & \multicolumn{1}{c|}{51.2} & 54.9
& 46.1 & 70.5 & 45.6 & 61.1 & \multicolumn{1}{c|}{51.2} & 54.9 & \textcolor{red}{-0.1}  & \textbf{54.72M}\\ 

\multicolumn{1}{l|}{} & \multicolumn{1}{l|}{TENT~\cite{tent}} & \textit{\textcolor{crimson}{Full}} &  38.3 & 60.9 & 36.6 & 48.3 & \multicolumn{1}{c|}{41.3} & 45.0 
& 35.8 & 57.6 & 33.6 & 44.3 & \multicolumn{1}{c|}{38.8} & 42.0 & \textcolor{red}{-10.8} & 0.02M \\ 

\multicolumn{1}{l|}{\textit{w/ SDA}} & \multicolumn{1}{l|}{SAR~\cite{sar}}  & \textit{\textcolor{crimson}{Full}} & 43.6 & 67.9 & 43.1 & 58.6 & \multicolumn{1}{c|}{48.5} 
& 52.3 & 43.4 & 67.4 & 42.2 & 58.1 & \multicolumn{1}{c|}{47.6} & 51.9 & \textcolor{red}{-1.0} & 0.02M \\ 

\multicolumn{1}{l|}{} & \multicolumn{1}{l|}{EcoTTA~\cite{ecotta}} & \textit{\textcolor{darkcerulean}{MetaNet}} & 42.3 & 67.3 & 32.1 & 55.8 & \multicolumn{1}{c|}{46.7} & 48.8 
& 41.1 & 65.6 & 27.0 & 53.2 & \multicolumn{1}{c|}{45.3} & 46.4 & \textcolor{red}{-6.5} & 3.46M\\ 


\multicolumn{1}{l|}{} & \multicolumn{1}{l|}{\cellcolor{g}BECoTTA+ (S)} &
\cellcolor{g}\textit{\textcolor{darkcerulean}{MoDE}} & 
\cellcolor{g}44.0 &
\cellcolor{g}69.4 &
\cellcolor{g}40.2 &
\cellcolor{g}56.8 &
\multicolumn{1}{c|}{\cellcolor{g}49.2} &
\cellcolor{g}51.9 &
\cellcolor{g}44.0 &
\cellcolor{g}69.4 &
\cellcolor{g}40.1 &
\cellcolor{g}56.9 &
\multicolumn{1}{c|}{\cellcolor{g}49.1} &
\cellcolor{g}51.9 &
\cellcolor{g}+0.0 & 
\cellcolor{g}0.12M \\ 

\multicolumn{1}{l|}{} & \multicolumn{1}{l|}{\cellcolor{gg}BECoTTA+ (M) } &
\cellcolor{gg}\textit{\textcolor{darkcerulean}{MoDE}} & 
\cellcolor{gg}45.6 & 
\cellcolor{gg}70.6 & 
\cellcolor{gg}42.6 & 
\cellcolor{gg}59.6 & 
\multicolumn{1}{c|}{\cellcolor{gg}\textbf{50.8}} & 
\cellcolor{gg}53.8 & 
\cellcolor{gg}45.6 & 
\cellcolor{gg}70.7 & 
\cellcolor{gg}42.5 & 
\cellcolor{gg}59.5 & 
\multicolumn{1}{c|}{\cellcolor{gg}\textbf{50.8}} & 
\cellcolor{gg}53.9 & 
\cellcolor{gg}+0.0 & 
\cellcolor{gg}0.77M \\ 

\multicolumn{1}{l|}{} & \multicolumn{1}{l|}{\cellcolor{ggg}BECoTTA+ (L) } &
\cellcolor{ggg}\textit{\textcolor{darkcerulean}{MoDE}} & 
\cellcolor{ggg}\textbf{45.6} &
\cellcolor{ggg}\textbf{71.3} &
\cellcolor{ggg}\textbf{43.7} &
\cellcolor{ggg}\textbf{59.5} &
\multicolumn{1}{c|}{\cellcolor{ggg}50.5} &
\cellcolor{ggg}\textbf{54.2} &
\cellcolor{ggg}\textbf{45.7} &
\cellcolor{ggg}\textbf{71.3} &
\cellcolor{ggg}\textbf{43.7} &
\cellcolor{ggg}\textbf{59.6} &
\multicolumn{1}{c|}{\cellcolor{ggg}50.6} &
\cellcolor{ggg}\textbf{54.2} &
\cellcolor{ggg}\textbf{+0.0} & 
\cellcolor{ggg}3.32M \\
\bottomrule
\end{tabular}%
}
\label{table:main2_ten}
\vspace{-0.1in}
\end{table*}


%% file: table/forgetting_metrics.tex
\begin{table}[t]
    \centering
    \scriptsize
    \caption{\textbf{Quantitative results of AvgIoU and BWT.} We evaluate AvgIoU and BWT among 3 rounds in the CDS-Hard scenario.}
    \renewcommand{\arraystretch}{1.6} 
    \begin{tabular}{l|cccccc|cc}
    \hline 
    \multicolumn{1}{c|}{} & \multicolumn{2}{c}{\textbf{Round 1}} & \multicolumn{2}{c}{\textbf{Round 2}} & \multicolumn{2}{c|}{\textbf{Round 3}} & \multicolumn{2}{c}{Avg} \\ \cline{2-9} 
    \multicolumn{1}{c|}{} & AvgIoU & BWT & AvgIoU & BWT & AvgIoU & BWT & AvgIoU & BWT \\ \hline  \hline
    TENT~\cite{tent}& 47.87 & -0.15 & 46.63 & -0.73 & 44.94 & -0.97 & 46.48 & -0.62 \\
    SAR~\cite{sar} & 48.11 & 0.08 & 48.19 & 0.04 & 48.23 & 0.01 & 48.18 & 0.04 \\ \hline
    
    \rowcolor[HTML]{EFEFEF} 
    BECoTTA (Ours) - M & 51.29 & 0.15 & 51.33 & 0.17 & 51.33 & 0.16 & 51.32 & 0.16 \\
    \rowcolor[HTML]{EFEFEF} 
    ~~~~~~~~~~~~~~~~\textit{+ SDA} & \textbf{54.32} & \textbf{0.26} & \textbf{54.30} & \textbf{0.30} & \textbf{54.30} & \textbf{0.31} & \textbf{54.31} & \textbf{0.29} \\ \hline
    \end{tabular}%
    \label{table:cl_metrics_v2}
\end{table}

%% file: table/full_hiddendim_ablation.tex
\begin{table*}[t]
\centering
\renewcommand{\arraystretch}{1.3}
\caption{\textbf{Further ablation study for the number of experts $N$, $K$, and hidden dimensions}. We report sufficient ablation studies about the $N$, $K$, and hidden dimensions in the MoDE layer. All experiments are conducted in \textit{w/ WAD} setting. '\textit{Last}' denotes the MoDE layer located in the last stage of the encoder, whereas '\textit{All}' denotes those located in every four stages of the encoder.}  
\resizebox{\columnwidth}{!}{%
{\scriptsize
\begin{tabular}{cc|ccc|cccccc}
\hline
 &  & \multicolumn{1}{c|}{} &  &  & \multicolumn{6}{c}{Round 1} \\
\multirow{-2}{*}{Parameters} & \multirow{-2}{*}{Memory} & \multicolumn{1}{c|}{\multirow{-2}{*}{Mode}} & \multirow{-2}{*}{Expert, K} & \multirow{-2}{*}{Hidden dim} & B-Clear & A-Fog & A-Night & A-Snow & \multicolumn{1}{c|}{B-Over} & Avg \\ \hline
\cellcolor[HTML]{FFFFFF}59,922 & 227.27MB & \multicolumn{1}{c|}{\cellcolor[HTML]{FFFFFF}} & \cellcolor[HTML]{FFFFFF}exp3 k1 & \cellcolor[HTML]{FFFFFF}[0, 0, 0, 2] & \cellcolor[HTML]{FFFFFF}44.10 & \cellcolor[HTML]{FFFFFF}69.46 & \cellcolor[HTML]{FFFFFF}39.13 & \cellcolor[HTML]{FFFFFF}57.24 & \multicolumn{1}{c|}{\cellcolor[HTML]{FFFFFF}49.13} & \cellcolor[HTML]{FFFFFF}51.81 \\
\cellcolor[HTML]{FFFFFF}129,096 & 227.80MB & \multicolumn{1}{c|}{\cellcolor[HTML]{FFFFFF}} & \cellcolor[HTML]{FFFFFF}exp4 k1 & \cellcolor[HTML]{FFFFFF}[0, 0, 0, 6] & \cellcolor[HTML]{FFFFFF}44.06 & \cellcolor[HTML]{FFFFFF}69.4 & \cellcolor[HTML]{FFFFFF}40.10 & \cellcolor[HTML]{FFFFFF}56.84 & \multicolumn{1}{c|}{\cellcolor[HTML]{FFFFFF}49.17} & \cellcolor[HTML]{FFFFFF}51.91 \\
\rowcolor[HTML]{FFFFFF} 
378,144 & 229.71MB & \multicolumn{1}{c|}{\cellcolor[HTML]{FFFFFF}} & exp6 k3 & [0, 0, 0, 16] & 44.31 & 69.15 & 40.10 & 57.52 & \multicolumn{1}{c|}{\cellcolor[HTML]{FFFFFF}49.64} & 52.14 \\
\cellcolor[HTML]{FFFFFF}1,208,448 & 236.46MB & \multicolumn{1}{c|}{\cellcolor[HTML]{FFFFFF}} & \cellcolor[HTML]{FFFFFF}exp6 k3 & \cellcolor[HTML]{FFFFFF}[0, 0, 0, 64] & \cellcolor[HTML]{FFFFFF}44.12 & \cellcolor[HTML]{FFFFFF}69.19 & \cellcolor[HTML]{FFFFFF}40.21 & \cellcolor[HTML]{FFFFFF}57.19 & \multicolumn{1}{c|}{\cellcolor[HTML]{FFFFFF}49.51} & \cellcolor[HTML]{FFFFFF}52.04 \\
\cellcolor[HTML]{FFFFFF}4,212,480 & 258.97MB & \multicolumn{1}{c|}{\cellcolor[HTML]{FFFFFF}} & \cellcolor[HTML]{FFFFFF}exp20 k10 & \cellcolor[HTML]{FFFFFF}[0, 0, 0, 64] & \cellcolor[HTML]{FFFFFF}44.21 & \cellcolor[HTML]{FFFFFF}69.10 & \cellcolor[HTML]{FFFFFF}40.20 & \cellcolor[HTML]{FFFFFF}57.30 & \multicolumn{1}{c|}{\cellcolor[HTML]{FFFFFF}49.50} & \cellcolor[HTML]{FFFFFF}52.06 \\
\rowcolor[HTML]{FFFFFF} 
4,074,240 & 242.89MB & \multicolumn{1}{c|}{\multirow{-6}{*}{\cellcolor[HTML]{FFFFFF}Last}} & exp10 k4 & [0, 0, 0, 128] & 43.91 & 68.48 & 39.31 & 56.52 & \multicolumn{1}{c|}{\cellcolor[HTML]{FFFFFF}49.28} & 51.50 \\
\rowcolor[HTML]{FFFFFF} \hline
779,916 & 231.36MB & \multicolumn{1}{c|}{\cellcolor[HTML]{FFFFFF}} & exp6 k3 & [2, 4, 10, 16] & 45.45 & 70.77 & 42.62 & 59.66 & \multicolumn{1}{c|}{\cellcolor[HTML]{FFFFFF}50.76} & 53.85 \\
\rowcolor[HTML]{FFFFFF} 
956,976 & 232.71MB & \multicolumn{1}{c|}{\cellcolor[HTML]{FFFFFF}} & exp6 k3 & [8, 8, 16, 16] & 45.54 & 70.40 & 42.76 & 59.50 & \multicolumn{1}{c|}{\cellcolor[HTML]{FFFFFF}50.90} & 53.82 \\
\rowcolor[HTML]{FFFFFF} 
1,252,176 & 234.96MB & \multicolumn{1}{c|}{\cellcolor[HTML]{FFFFFF}} & exp6 k3 & [8, 8, 16, 32] & 45.21 & 70.53 & 43.26 & 59.32 & \multicolumn{1}{c|}{\cellcolor[HTML]{FFFFFF}50.69} & 53.80 \\
\rowcolor[HTML]{FFFFFF} 
1,299,860 & 236.92MB & \multicolumn{1}{c|}{\cellcolor[HTML]{FFFFFF}} & exp10 k4 & [2, 4, 10, 16] & 45.47 & 69.56 & 43.04 & 58.88 & \multicolumn{1}{c|}{\cellcolor[HTML]{FFFFFF}50.54} & 53.50 \\
\rowcolor[HTML]{FFFFFF} 
2,599,720 & 247.03MB & \multicolumn{1}{c|}{\multirow{-5}{*}{\cellcolor[HTML]{FFFFFF}All}} & exp20 k10 & [2, 4, 10, 16] & 45.33 & 70.17 & 43.13 & 59.48 & \multicolumn{1}{c|}{\cellcolor[HTML]{FFFFFF}51.04} & 53.83 \\ \hline
\rowcolor[HTML]{EFEFEF} 
3,469,312 & 251.21MB & \multicolumn{3}{c|}{\cellcolor[HTML]{EFEFEF}EcoTTA + ViT (w/WAD)} & 44.64 & 70.21 & 41.68 & 58.02 & \multicolumn{1}{c|}{\cellcolor[HTML]{EFEFEF}49.94} & 52.9 \\ \hline
\rowcolor[HTML]{FFFFFF} 
4,554,528 & 261.60MB & \multicolumn{1}{c|}{\cellcolor[HTML]{FFFFFF}} & exp3 k1 & [32, 64, 160, 256] & 45.28 & 70.22 & 43.14 & 58.74 & \multicolumn{1}{c|}{\cellcolor[HTML]{FFFFFF}50.42} & 53.56 \\
\rowcolor[HTML]{FFFFFF} 
6,072,704 & 273.20MB & \multicolumn{1}{c|}{\cellcolor[HTML]{FFFFFF}} & exp4 k3 & [32, 64, 160, 256] & 45.16 & 70.41 & 42.91 & 59.65 & \multicolumn{1}{c|}{\cellcolor[HTML]{FFFFFF}50.39} & 53.70 \\
\rowcolor[HTML]{FFFFFF} 
9,109,056 & 296.39MB & \multicolumn{1}{c|}{\cellcolor[HTML]{FFFFFF}} & exp6 k3 & [32, 64, 160, 256] & 45.45 & 70.61 & \textbf{44.32} & 59.40 & \multicolumn{1}{c|}{\cellcolor[HTML]{FFFFFF}50.61} & 54.08 \\ 
\rowcolor[HTML]{FFFFFF} 
15,181,760 & 342.77MB & \multicolumn{1}{c|}{\multirow{-4}{*}{\cellcolor[HTML]{FFFFFF}All}} & exp10 k4 & [32, 64, 160, 256] & \textbf{46.17} & \textbf{71.3} & 43.47 & \textbf{60.63} & \multicolumn{1}{c|}{\cellcolor[HTML]{FFFFFF}\textbf{51.18}} & \textbf{54.55} \\ \hline
\rowcolor[HTML]{EFEFEF} 
477,805,276 & 533.81MB & \multicolumn{3}{c|}{\cellcolor[HTML]{EFEFEF}Scratch CoTTA (w/WAD)} & 46.42 & 70.64 & 45.7 & 61.2 & \multicolumn{1}{c|}{\cellcolor[HTML]{EFEFEF}51.32} & 55.06 \\ \hline
\end{tabular}%
}
}
\label{table:hiddendim}
\end{table*}